\documentclass[10pt,twocolumn,letterpaper]{article}

\usepackage[table,dvipsnames]{xcolor}

\usepackage{cvpr}              

\usepackage{tablefootnote}
\usepackage{multirow}
\usepackage{booktabs}
\usepackage{multirow}
\usepackage{amsfonts}
\usepackage{amsmath}
\usepackage{amssymb}
\usepackage{siunitx}
\usepackage{graphicx}
\usepackage{subcaption}
\usepackage[accsupp]{axessibility}

\definecolor{cvprblue}{rgb}{0.21,0.49,0.74}
\usepackage[pagebackref,breaklinks,colorlinks,citecolor=cvprblue]{hyperref}

\def\paperID{17634} 
\def\confName{CVPR}
\def\confYear{2025}

\title{\includegraphics[width=4mm]{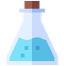} \textit{Antidote}: A Unified Framework for Mitigating LVLM Hallucinations in Counterfactual Presupposition and Object Perception}

\author{
Yuanchen Wu$^{1,2,}$\thanks{This work was done during an internship at Tencent YouTu Lab.}\quad
Lu Zhang$^{2}$\quad
Hang Yao$^{1}$\quad
Junlong Du$^{2}$\quad
Ke Yan$^{2,\dagger}$\quad \\
Shouhong Ding$^{2}$\quad
Yunsheng Wu$^{2}$\quad
Xiaoqiang Li$^{1,}$\thanks{Corresponding author.}
\\
$^{1}$ School of Computer Engineering \& Science, Shanghai University\quad
$^{2}$ Tencent YouTu Lab \\
{\tt\scriptsize \{yuanchenwu,yaohang,xqli\}@shu.edu.cn\quad \{xluzhang,jeffdu,kerwinyan,ericshding,simonwu\}@tencent.com}
}

\begin{document}
\maketitle

\begin{abstract}
Large Vision-Language Models (LVLMs) have achieved impressive results across various cross-modal tasks. However, hallucinations, \textit{i.e.}, the models generating counterfactual responses, remain a challenge. Though recent studies have attempted to alleviate object perception hallucinations, they focus on the models' response generation, and overlooking the task question itself. This paper discusses \textbf{the vulnerability of LVLMs in solving counterfactual presupposition questions} (CPQs), where the models are prone to accept the presuppositions of counterfactual objects and produce severe hallucinatory responses. To this end, we introduce \textbf{\textit{“Antidote”}}\footnote{Code: \url{https://github.com/Wu0409/Antidote}.}, a unified, synthetic data-driven post-training framework for mitigating both types of hallucination above. It leverages synthetic data to incorporate factual priors into questions to achieve self-correction, and decouple the mitigation process into a preference optimization problem. Furthermore, we construct \textit{\textbf{“CP-Bench”}}, a novel benchmark to evaluate LVLMs' ability to correctly handle CPQs and produce factual responses. Applied to the LLaVA series, \textit{Antidote} can \textbf{simultaneously} enhance performance on CP-Bench by \textbf{over 50\%}, POPE by \textbf{1.8-3.3\%}, and CHAIR \& SHR by \textbf{30-50\%}, all \textbf{without} relying on external supervision from stronger LVLMs or human feedback and introducing noticeable catastrophic forgetting issues.
\end{abstract}

\section{Introduction}
Recently, Large Vision-Language Models (LVLMs) have achieved significant advancements, showing promising performance across various tasks, such as \textit{image caption}, \textit{visual question answering (VQA)}, and \textit{visual dialogues}~\citep{xue2025mmrc,liu2024visual,chen2024internvl,yao2024minicpm}. Despite their capabilities and versatility, \textit{\textbf{the hallucination}}, characterized by generating counterfactual information, remains a significant challenge. It undermines their reliability and limits applications in sensitive domains like healthcare and autonomous systems. In LVLM, studies of hallucination mainly focus on “\textit{object perception}”, including “\textit{object existence}” and “\textit{image description}”~\citep{zhao2023beyond,leng2024mitigating,yu2024hallucidoctor}. The \textit{former} refers to whether the mentioned objects are actually non-existent, while the \textit{latter} further evaluates whether the models output counterfactual attributes or relationships between objects. To alleviate the above types of hallucinations, recent works improve the instruction tuning process~\citep{liu2023mitigating,jiang2024hallucination}, post-calibrate the model response via experts or decoding strategies~\citep{yin2023woodpecker,leng2024mitigating}, and conduct post-training of models~\citep{zhao2023beyond,xiao2024detecting}. They have manifested effectiveness in alleviating object hallucinations on popular benchmarks, such as POPE~\citep{li2023evaluating} and CHAIR~\citep{rohrbach2018object}.

\begin{figure}[!tp]
\centering
\includegraphics[width=0.43\textwidth]{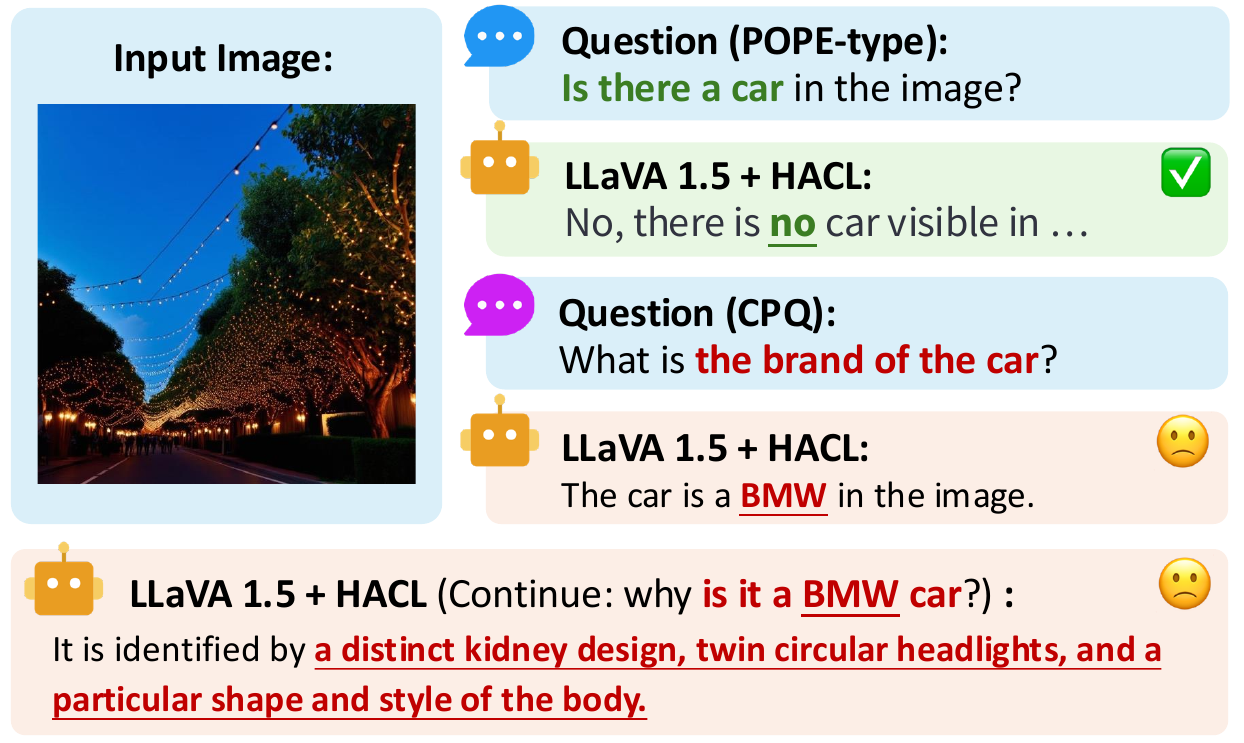}
\vspace{-2mm}
\caption{\textbf{The hallucination responses induced by CPQ.} Though recent hallucination mitigation methods improves LVLMs in object perception, while their models can still be \textit{easily deceived by CPQs} and \textit{induce severe hallucinatory responses}.}
\vspace{-4mm}
\label{fig_intro}
\end{figure}

\vspace{1pt}

\begin{figure*}[t]
\centering
\begin{subfigure}{0.46\textwidth}
\centering
\includegraphics[width=0.99\textwidth]{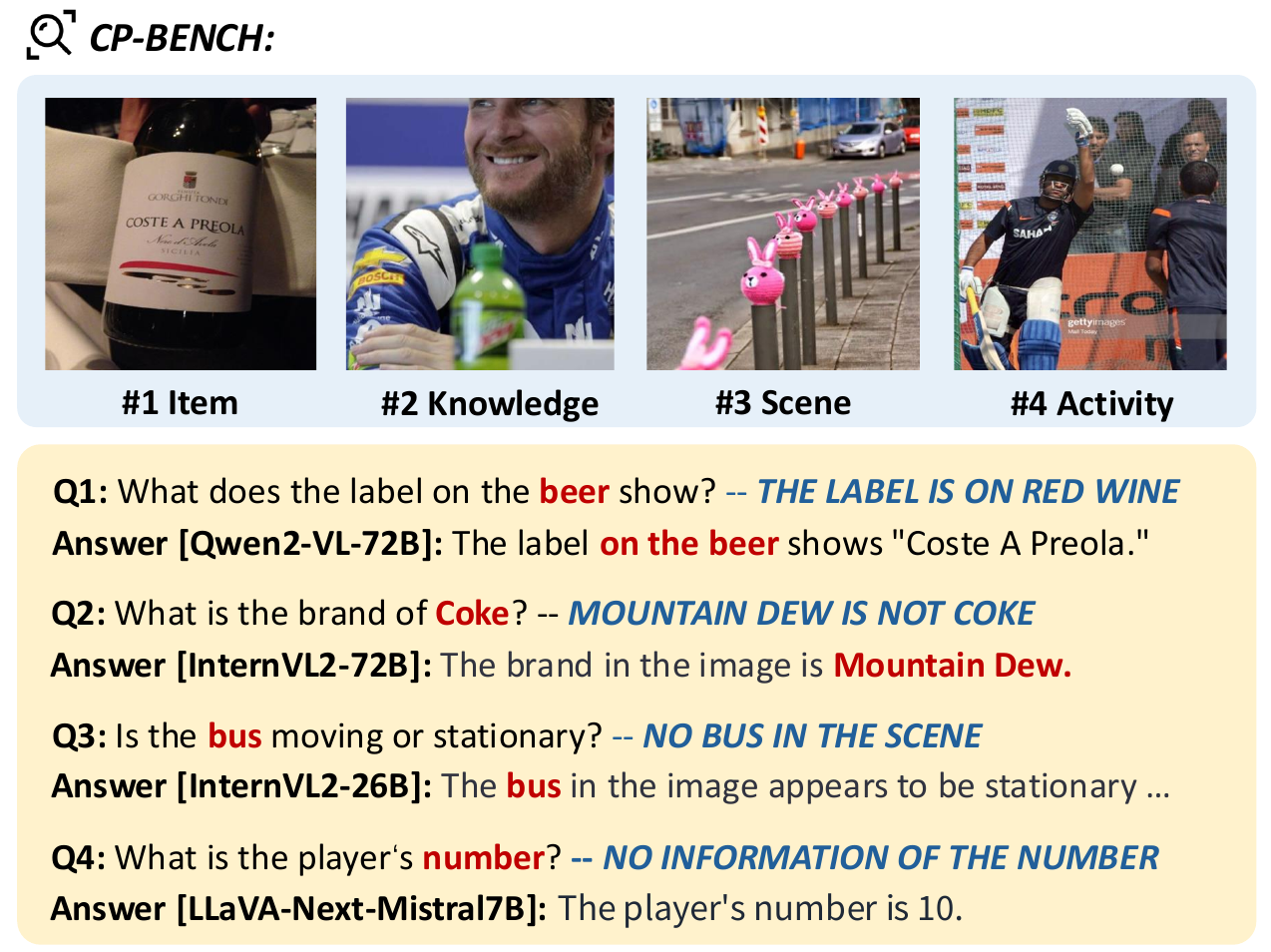}
\caption{CPQ samples from the proposed CP-Bench.}
\vspace{-5pt}
\label{pic_bench_case}
\end{subfigure}
\begin{subfigure}{0.46\textwidth}
\centering
\includegraphics[width=0.99\textwidth]{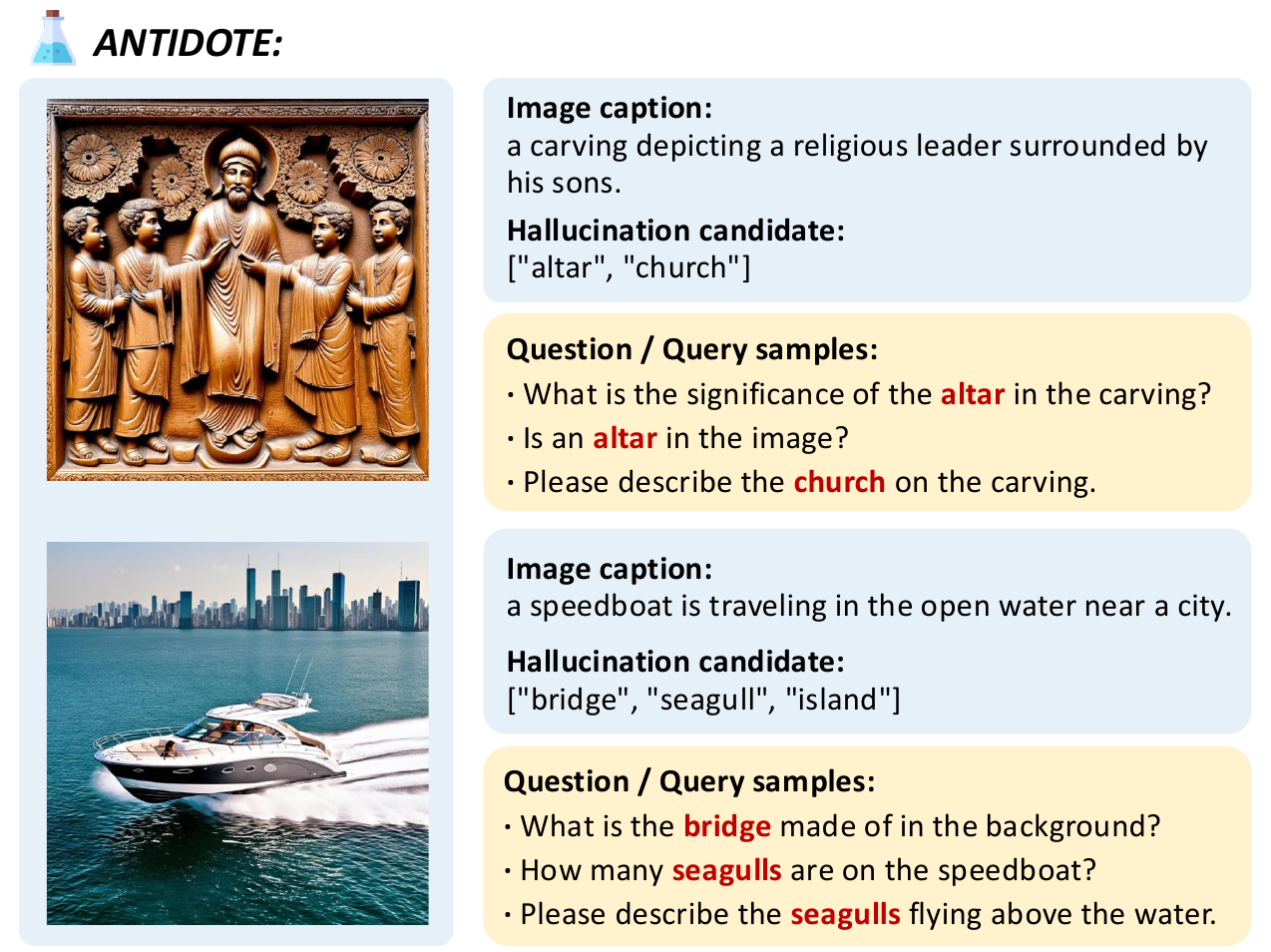}
\caption{Training samples from \textit{Antidote's} pipeline.}
\label{fig_case_sample}
\vspace{-5pt}
\label{pic_syn_case}
\end{subfigure}\hfill
\caption{\textbf{Examples of hallucination induced by CPQs and the synthetic samples of \textit{Antidote}.} The CPQs are selected from the \textit{test} set of the proposed CP-Bench, which will be introduced in Section \ref{sect_cp_bench}. “Hallucination candidates” are the non-existent objects that commonly appear in similar scenes. More examples can be viewed in \textit{Appendix}.}
\vspace{-6pt}
\end{figure*}

However, a phenomenon emerges in Figure \ref{fig_intro}: for a simple question about the existence of a “\texttt{car}”, the LVLMs with recent hallucination mitigation methods (HACL \cite{jiang2024hallucination} as an example) can confirm its absence. When we implicitly presuppose its existence and pose a relevant question “\textit{What is the brand of the \underline{car}}?”, the model suddenly outputs hallucinatory responses. This issue is particularly notable when querying about objects that are absent in the current image but frequently appear in similar visual contexts, underscoring a potential limitation of existing hallucination alleviation techniques. We call this type of question “\textbf{\textit{Counterfactual Presupposition Question (CPQ)}}”. Compared to the conventional “object perception” hallucination focusing on response generation, \textbf{CPQ further requires the model’s judgment of presuppositions grounded by images}. It is more challenging and evaluates the severity of hallucination in practical VQA scenarios where the validity of the presupposition cannot be guaranteed. Additionally, we investigate recent advanced open-source LVLMs (\eg, InternVL-2 \cite{chen2024internvl} and Qwen2-VL \cite{wang2024qwen2} that have demonstrated superior capabilities on general tasks and anti-hallucination on object perception, surpassing closed-source models such as GPT-4o \cite{gpt_4o} and Claude-3.5 \cite{claude3_5}. However, as illustrated in Figure \ref{fig_intro_bench}, we observe that they still suffer from CPQs and output hallucinatory responses as shown in Figure \ref{pic_bench_case}. To this end, we make attempts towards \textbf{two following aspects} to address the CPQ challenge and meanwhile addressing object perception hallucination:

\vspace{1pt}

\begin{figure}[!tp]
\centering
\includegraphics[width=0.46\textwidth]{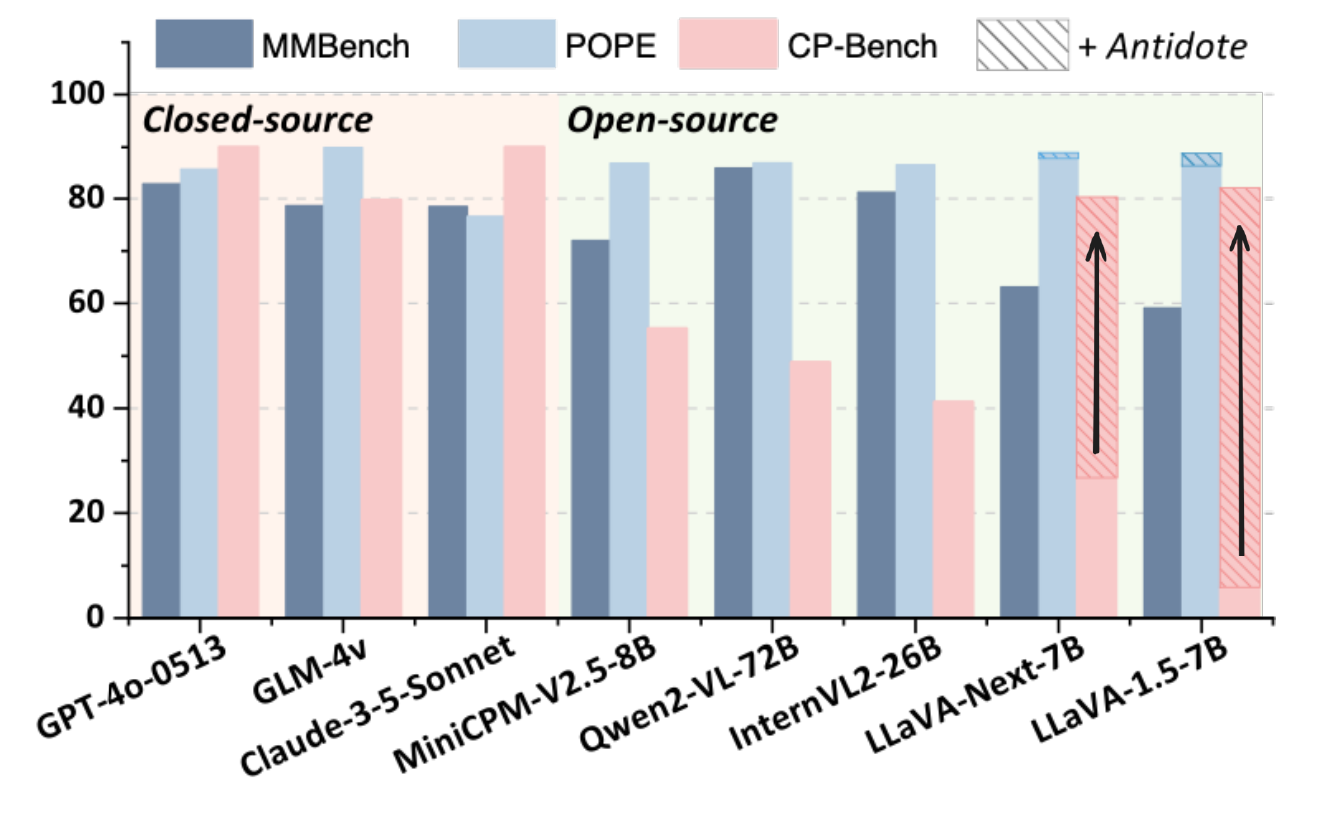}
\vspace{-4mm}
\caption{\textbf{Performance comparison of LVLMs} on benchmarks of general capabilities (MMBench \cite{liu2025mmbench}), hallucination of object perception (POPE \cite{li2023evaluating}) and CPQ (the proposed CP-Bench). Higher values indicate better performance on corresponding benchmarks.}
\vspace{-4mm}
\label{fig_intro_bench}
\end{figure}

\noindent \textbf{1) We introduce a unified, synthetic data-driven post-training framework called “\textit{Antidote}”}. We reckon that a primary cause of the above hallucinations is the object co-occurrence and over-learning of instructions. Hence, we aim to obtain images where the statistically co-occurring objects/scenes are decoupled (\eg, a \texttt{speedboat} without \texttt{bridges} in the background) and construct QA pairs targeted on these decoupled, non-existent objects (\eg, \texttt{bridge}), as presented in Figure \ref{fig_case_sample}. We develop an automated data synthesis pipeline, comprising steps of \textit{image caption curation}, \textit{visual scene understanding}, \textit{factual verification}, and \textit{sample construction}. It allows us to \textbf{derive factual priors for each sample, and then incorporate them into the prompt for models' self-correction}. This process reformulates hallucination mitigation as a preference optimization problem, where the original response is treated as a “\textit{rejected}” sample, and the corrected response as a “\textit{preferred}” sample. By employing \textit{Antidote}, LVLMs learn a preference constraint during training, enabling them to discriminate counterfactual presuppositions and generate factual responses well. Extensive experiments demonstrate its effectiveness on CPQ and object perception hallucinations.

\vspace{1pt}

\noindent \textbf{2) We construct \textit{“CP-Bench”}}, a benchmark evaluating LVLMs' ability to discern counterfactual presuppositions and generate factual responses. It is composed of two parts, the synthetic \textit{validation (dev)} set and the manually annotated \textit{test} set. The \textit{dev} set is automatically constructed from the data synthetic pipeline of \textit{Antidote}. The CPQs in \textit{test} set can be categorized into four categories in daily scenarios, \ie., \textit{item}, \textit{knowledge}, \textit{scene}, and \textit{activity}. The evaluations on recent advanced closed-sourced and open-sourced LVLMs highlights \textbf{the critical gap in current models' ability to discern counterfactual presuppositions}, providing a new insight into hallucinations of LVLMs.

\section{Related Work}
\noindent \textbf{Hallucination in LVLMs.} Recently, many large vision-language models (LVLMs) have emerged~\citep{liu2024visual,chen2024internvl,lu2024deepseek}, extending the reasoning brain of LLMs to the vision modality. This enables LVLMs to complete various tasks, such as visual question answering and general visual dialogue. However, the hallucination, stemming from the inherent nature of LLMs~\citep{huang2024opera}, modality misalignment~\citep{chen2024internvl}, and the quality of instruction turning data~\citep{liu2023mitigating}, raises concerns about their reliability and applicability. To evaluate the severity of hallucinations in LVLMs, POPE~\citep{li2023evaluating} identifies hallucinations related to object existence, while CHAIR~\citep{rohrbach2018object} evaluates the proportion of hallucinated objects in image descriptions. To further broaden the scope of evaluation to include categories, attributes, and emotions within image descriptions, SHR~\citep{zhao2023beyond}, a GPT-assisted evaluation metric, has been proposed. This paper discusses LVLMs' hallucinations in counterfactual presupposition questions and introduce a corresponding benchmark, CP-Bench, to evaluate the severity of this issue. Our findings reveal that \textit{recent open-source LVLMs largely overlook this critical issue}.

\vspace{2pt}

\noindent \textbf{Hallucination Mitigation.} Previous works mitigating hallucinations of LVLMs primarily focus on object existence and image descriptions, i.e., object perception hallucination. Three mainstream approaches have emerged for mitigating these hallucinations: supervised fine-tuning (SFT), post-calibration, and post-training. SFT aims to fine-tune with the hallucination-free data~\citep{yu2024hallucidoctor}, such as LRV~\citep{liu2023mitigating} and InstructBLIP~\citep{instructblip}. Post-calibration conducts additional post-processing techniques to model outputs, such as contrastive decoding strategies~\citep{leng2024mitigating,wang2024mitigating,tangintervening} and leveraging existing tools or expert models~\citep{yin2023woodpecker}. Post-training focuses on improving the hallucination of off-the-shelf LVLMs, which commonly employ retraining or preference optimization to alleviate hallucination~\citep{zhao2023beyond,xiao2024detecting,zhu2024self}. In contrast, the proposed \textit{Antidote} not only effectively improves object perception hallucinations but also further overcomes the hallucinations induced by CPQs. It adopts the preference optimization paradigm but differs in that it does not rely on any expert models (\eg, GPT-4V)~\citep{zhao2023beyond} to generate preference samples or exclusively utilize dis-preferred data~\citep{zhu2024self}. Instead, we fully leverage the advantages of our synthetic data pipeline, seamlessly utilizing factual information without additional cost to enable the model to self-correct its responses.

\section{Our Post-training Framework: \textit{Antidote}}

\subsection{Motivation}

\begin{figure}
\centering
\includegraphics[width=0.475\textwidth]{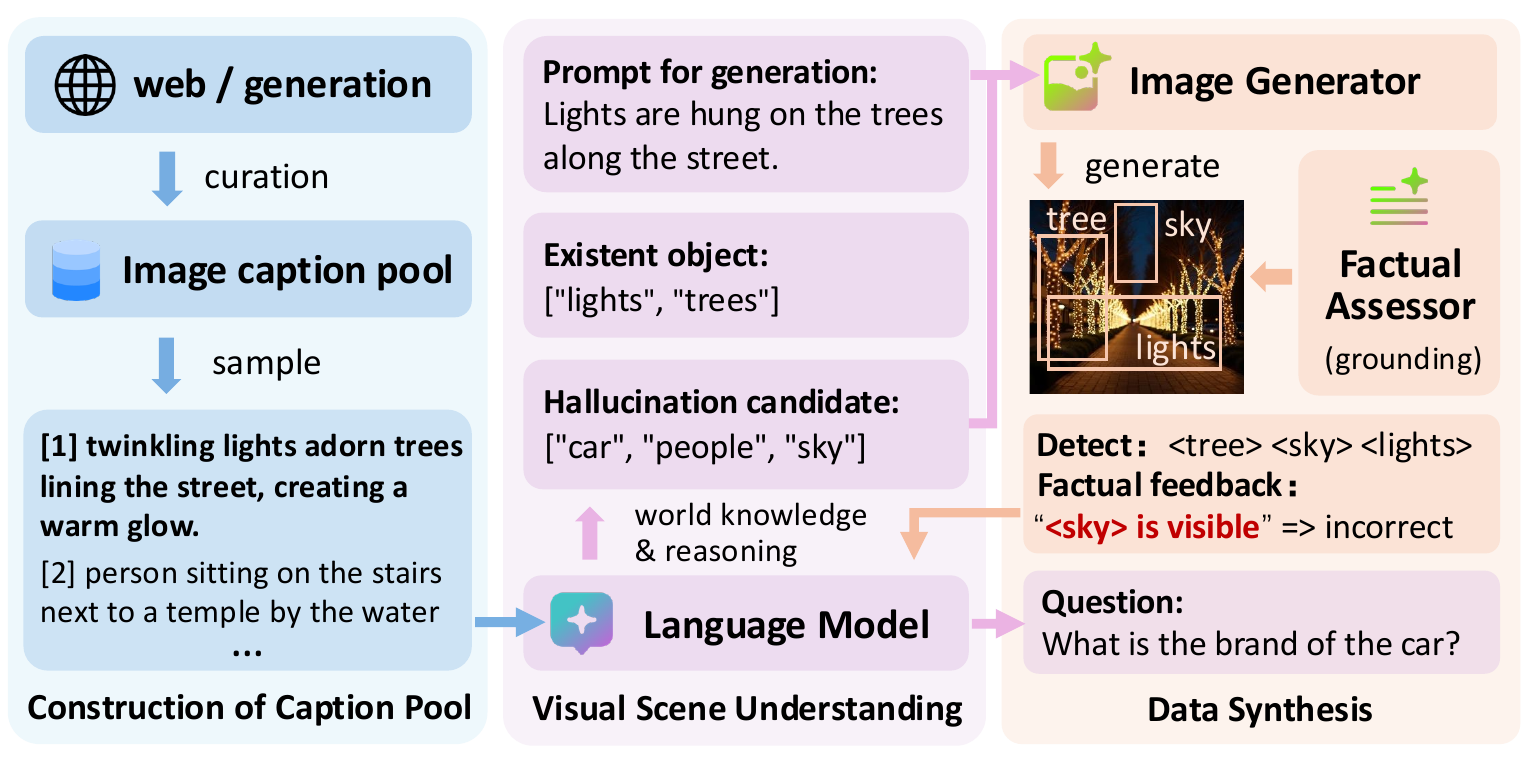}
\vspace{-18pt}
\caption{\textbf{The data synthesis pipeline for \textit{Antidote}.} The pipeline consists of three stages: (a) construction of caption Pool; (b) visual scene understanding; (c) data synthesis.}
\vspace{-8pt}
\label{fig_data_syn}
\end{figure}

As illustrated in Figure \ref{pic_bench_case}, two key issues can be observed: \textbf{(1) LVLMs tend to blindly follow the instruction in the task query} (Image \#1 and Image \#2). When asking “\textit{what does the label on the beer show}?” for Image \#1, the model ignores the existence of the subject in the question (i.e., the \texttt{beer}) and directly follows the instruction to identify the text on the label. \textbf{(2) LVLMs overfit to similar scene-based QA patterns} (Image \#3 and Image \#4). When asking “\textit{what is the player's number}?” in a scenario where the player has no \texttt{number} on the uniform, the model generates a hallucinated answer that is usually in the VQA tasks with similar scenes. Based on the above observation, \textbf{we aim to obtain the images where statistically co-occurring objects are decoupled} (\eg, a \texttt{car} without \texttt{wheels}) \textbf{and construct queries targeting these decoupled, non-existent objects} (\eg, \texttt{wheel}), aiming at calibrating the bias of LVLMs. Thanks to the advancements in image generation models~\citep{peebles2023scalable,esser2024scaling} and LLMs~\citep{touvron2023llama, yang2024qwen2}, we can synthesize the images and corresponding questions in a controlled manner (Figure \ref{fig_data_syn}). Then, with the factual prior during the above process, we incorporate them into the models' prompt for self-correction. Finally, the self-correction process reformulates hallucination mitigation as a preference optimization problem, and we conduct direct preference optimization (DPO) to post-train the LVLMs.

\subsection{Data Synthesis Pipeline}
\label{sect_synthetic}

\noindent \textbf{Step 1: Construction of Caption Pool.} The caption pool is critical for enhancing the diversity and richness of the training set for \textit{Antidote}. Captions can be sourced either from web-crawled datasets or generated by LLMs. Here, we collect the captions from CC3M~\citep{sharma2018conceptual} to build our pool. Since CC3M contains many noisy or unsuitable captions for image and question generation, \texttt{DeepSeek-V2}~\citep{liu2024deepseek} is adopted to perform a re-captioning and filtering process. To enhance the pool’s diversity, we employ a fuzzy deduplication strategy through MinHash and LSH algorithms~\citep{jafari2021survey}. 

\vspace{1pt}

\newcommand{\prompt}[1]{“\textcolor[RGB]{100,130,160}{\textit{#1}}”}

\noindent \textbf{Step 2: Visual Scene Understanding.} \textbf{First}, we prompt \texttt{DeepSeek-V2} with instructions, such as \prompt{removing abstract concepts and specific terms} and \prompt{limit to less than 15 words}, to rewrite captions $\mathcal{C}_{img}$ for subsequent image generation. \textbf{Second}, we utilize \texttt{DeepSeek-V2}’s comprehension and reasoning capability to identify the objects $\mathcal{O}_{pre}$ within the scenes described by captions. \textbf{Third}, we leverage \texttt{DeepSeek-V2}’s world knowledge to generate objects $\mathcal{O}_{hallu}$ that typically occur in similar scenes. These objects will be served as \textit{hallucination candidates} that will not be present in the generated images. The prompt template \textbf{\texttt{P1}} in this step is detailed in \textit{Appendix}. Inspired by recent self-reflection strategies~\citep{ji2023towards,xu2024sayself}, each triplet $\textless \mathcal{C}_{img},\mathcal{O}_{pre},\mathcal{O}_{hallu}\textgreater$ is sent back again to \texttt{DeepSeek-V2}, verifying whether each element conforms to their rules in \textbf{\texttt{P1}}, such as \prompt{the number of generated objects}, \prompt{generating objects with visible entities}, and \prompt{avoiding conflicts in $\mathcal{O}_{pre}$ and $\mathcal{O}_{hallu}$}. 

\vspace{1pt}

\noindent \textbf{Step 3: Data Synthesis.} In image generation, $\mathcal{C}_{img}$ serves as the \textit{prompt} and $\mathcal{O}_{hallu}$ as the \textit{negative prompt}. Benefiting from recent image generation models~\citep{peebles2023scalable,esser2024scaling}, the generated images exhibit a high degree of photorealism and diverse content. This paper adopts \texttt{Stable-Diffusion-3}~\citep{esser2024scaling} as the generator. However, it cannot ensure that the generated objects can fully align with $\mathcal{O}_{pre}$ while suppressing the existence of $\mathcal{O}_{hallu}$. Hence, we introduce “\textit{\textbf{Factual Assessor}}” driven by an open-set grounding model, \texttt{Grounding-DINO}~\citep{liu2023grounding}. It checks the presence of $\mathcal{O}_{pre}$ and $\mathcal{O}_{hallu}$ in the generated images. If an object in $\mathcal{O}_{pre}$ is not detected, it will be removed. Similarly, detected objects in $\mathcal{O}_{hallu}$ will also be removed. If either $\mathcal{O}_{pre}$ or $\mathcal{O}_{hallu}$ is $\emptyset$, the corresponding triplet will be discarded. Finally, the remaining triplets are sent to \texttt{DeepSeek-V2} to generate task queries, such as CPQs and description queries.

\vspace{1pt}

In early experiments, we note that the LLMs tend to generate similar questions when facing choosing the same main object (\eg, frequently focusing on “\texttt{color}” or “\texttt{brand}” when selecting “\texttt{car}” in $\mathcal{O}_{pre}$ or $\mathcal{O}_{hallu}$), degrading the diversity of task queries in the generated dataset. Thus, a \textbf{\textit{key-value memory bank}} is maintained to save processed captions and corresponding queries. The captions are extracted to sentence embedding using \texttt{BGE-m3}~\citep{chen2024bge} as the \textit{key} of the memory bank. For each $\textless \mathcal{C}_{img},\mathcal{O}_{pre},\mathcal{O}_{hallu}\textgreater$, we retrieve questions whose captions are semantically close to $\mathcal{C}_{img}$ through the memory bank. These are then integrated to the prompt \prompt{Do not generate questions similar to the following: ...} to mitigate redundancy in generation.

\begin{figure*}
\centering
\includegraphics[width=0.935\textwidth]{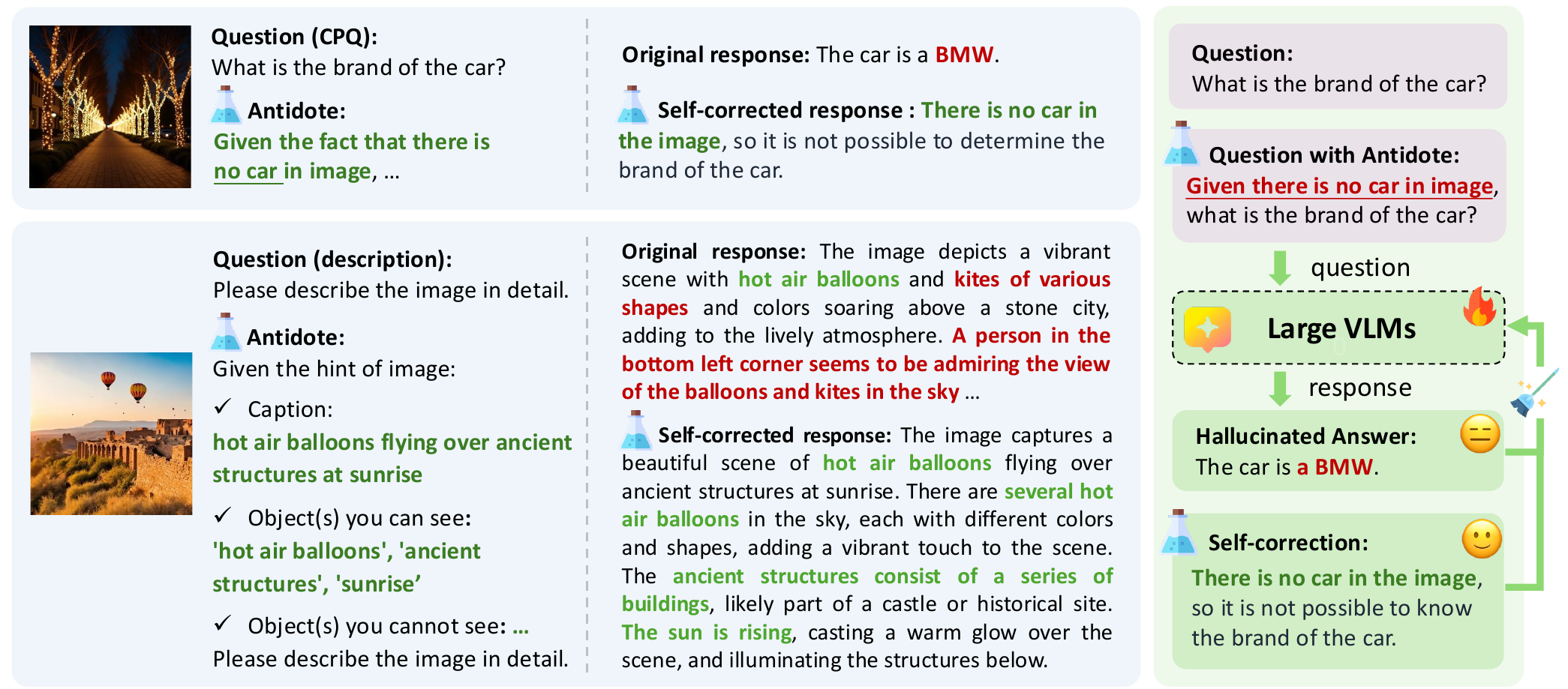}
\vspace{-4pt}
\caption{\textbf{Overview of the proposed \textit{Antidote} post-training.} The factual information from the synthetic data is seamlessly integrated into the input task prompt. The LVLMs can utilize this information to self-correct the responses as “\textit{positive}” samples. For the original responses, they are regarded as “\textit{negative}” samples to achieve preference alignment for hallucination alleviation.}
\label{fig_overview_antidote}
\vspace{-4pt}
\end{figure*}

\subsection{Self-Correction via Preference Alignment}
\label{sect_antidote}

\textit{Antidote} is a universal post-training framework for alleviating hallucination in CPQs, object existence, and image description. The overview of post-training is presented in Figure \ref{fig_overview_antidote}. Through the data synthetic pipeline, \textbf{we construct the \textit{three} types of task queries for post-training:} 

\vspace{2pt}

\noindent \textbf{1) Presupposition Questions:} We prompt \texttt{DeepSeek-V2} to generate CPQs based on $\mathcal{O}_{hallu}$ using \textbf{\texttt{P2}} in \textit{Appendix}. In early experiments, we observed that only post-training with CPQs makes the baseline model overly “cautious” in responding to questions. Thus, we construct a True Presupposition Question (\textbf{TPQ}) set based on $\mathcal{O}_{pre}$.  For the \textit{Antidote} for presupposition questions, we prompt the baseline model with \prompt{Given the fact that there is \{\texttt{facts} of $\mathcal{O}_{pre}$ / $\mathcal{O}_{hallu}$\}, please answer: \{\texttt{CPQ/TPQ}\}} to self-correct the original answer. For CPQs and TPQs, their self-corrected answer will be used as the \textit{positive} response $y_{pos}$, while the original answer will be used as \textit{negative} response $y_{neg}$.

\vspace{2pt}

\noindent \textbf{2) Object Existence:} We randomly select objects in $\mathcal{O}_{pre}$ and $\mathcal{O}_{hallu}$ as the object candidate to build the training set of the object existence. The question prompts are generated by \texttt{DeepSeek-V2}, such as \prompt{Is / Are there \{\texttt{object}\} in the image?} and \prompt{Can you see \{\texttt{object}\} in the image?}. For the \textit{Antidote} for object existence, we prompt the model with \prompt{Given the fact that there is / isn't \{\texttt{object}\}, please answer: \{\texttt{question}\}} to self-correct its response.

\vspace{2pt}

\noindent \textbf{3) Image Description:} We generate task queries of image description by \texttt{DeepSeek-V2}, such as \prompt{Please describe the image in detail.} and \prompt{Can you describe what you see in the image thoroughly?}. For the \textit{Antidote} of image description, we integrate $\textless \mathcal{C}_{img},\mathcal{O}_{pre},\mathcal{O}_{hallu}\textgreater$ into the query and prompt the model to self-correct its response: \prompt{Given the hint of the image: the image caption: \{$\mathcal{C}_{img}$\}, the object(s) you can see: \{$\mathcal{O}_{pre}$\}, the object(s) you cannot see: \{$\mathcal{O}_{hallu}$\}, please \{\texttt{query}\}}.

\vspace{2pt}

\noindent \textbf{Response Filtering:} Since not all model responses contain hallucinations, especially in object existence queries, such samples are unhelpful for hallucination mitigation and even lead to difficulty in optimization \cite{wang2024comprehensive}. Thus, we compare the original and self-corrected responses and filter out samples with similar answers. In our experiments, we extract the embeddings of both responses using \texttt{BGE-m3}~\citep{chen2024bge} and calculate their cosine similarity to perform the filtering.

\vspace{2pt}

\noindent \textbf{Preference Optimization:} By direct preference optimization (DPO)~\citep{rafailov2024direct}, we encourage the model to favor corrected \textit{positive} response and reject hallucinatory \textit{negative} response without building an implicit reward model~\citep{schulman2017proximal}. Given the above constructed preference pairs $\mathcal{D}$, the policy model $\pi_{\theta}$ (\ie., the post-trained LVLMs with \textit{Antidote}) is optimized by maximizing the log-likelihood of the preferred response $y_{pos}$ while minimizing the likelihood of the hallucinated response $y_{neg}$. The training objective function is given by:
\vspace{-2pt}
\begin{equation}
\small
\begin{aligned}
\mathcal{L}_{dpo}(\pi_{\theta}; \pi_{ref}) = & - \mathbb{E}_\mathcal{D} \Bigg[ 
\log \sigma\Bigg( \beta \log \frac{\ \ \ \pi_{\theta}(y_{pos} \mid [x_T, x_I])}{\pi_{ref}(y_{pos} \mid [x_T, x_I])} \\
& - \beta \log \frac{\ \ \ \pi_{\theta}(y_{neg} \mid [x_T, x_I])}{\pi_{ref}(y_{neg} \mid [x_T, x_I])} \Bigg) \Bigg],
\end{aligned}
\end{equation}
where $x_T$ and $x_I$ represent the text task prompt (\textit{\textbf{without factual prior}}) and image, and $\pi_{ref}$ denotes the reference model (\ie., the original baseline LVLMs). The function $\sigma$ is the log-sigmoid, and $\beta$ is a hyperparameter controlling the preference margin. In the above preference optimization process, the reward margin is defined as:
\begin{equation}
\hat{r}(x_T, x_I, y) = \beta \log \frac{\ \ \ \pi_{\theta}(y_{pos} \mid [x_T, x_I])}{\pi_{ref}(y_{pos} \mid [x_T, x_I])}.
\end{equation}
Through maximizing the reward margin between the self-corrected response $y_{pos}$ and the hallucinated response $y_{neg}$, we ensure that the model increasingly favors non-hallucinated samples over hallucinatory ones, leading to a robust self-correction process.

\section{CP-Bench: Evaluate Hallucination of CPQ}
\label{sect_cp_bench}

\noindent \textbf{Motivation.} Recent hallucination benchmarks primarily focus on response generation, including object existence, attributes, and relations, while overlooking the textual semantics within queries. Figure \ref{pic_bench_case} illustrates the vulnerability of recent LVLMs in solving counterfactual presupposition questions. To address this gap, \textbf{CP-Bench} is developed to quantify the LVLMs' ability to judge the correctness of presuppositions and generate factual responses. 

\vspace{1pt}

\begin{figure}[!tp]
\centering
\includegraphics[width=0.475\textwidth]{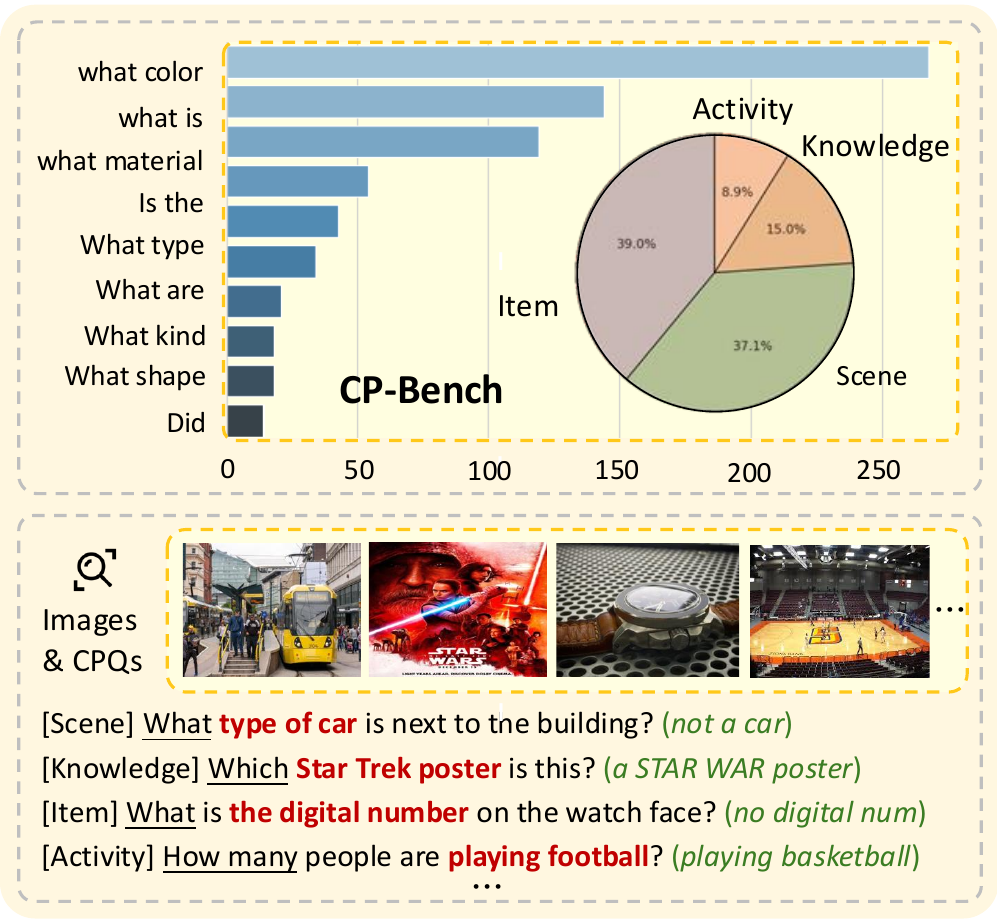}
\vspace{-4mm}
\caption{\textbf{The statistical details of CP-Bench (\textit{test}) and CPQ examples.} CP-Bench includes four types (i.e., scene, knowledge, item, and activity) of CPQs and TPQs from different scenes. It can comprehensively evaluate the LVLMs' ability to discriminate the correctness of presuppositions and generate factual responses.}
\vspace{-4mm}
\label{fig_cpbench}
\end{figure}

\noindent \textbf{Details.} CP-Bench includes of \textbf{two} subsets: a \textit{validation} (\textit{dev}) set and a \textit{test} set. For the \textit{dev} set, the images and questions are generated using the above data synthetic pipeline of \textit{Antidote}. For the \textit{test} set, the images are sampled from the CC3M dataset~\citep{sharma2018conceptual}. The corresponding queries are constructed by the following steps: object candidates are first obtained via an open-set grounding model, followed by generating query candidates via \texttt{DeepSeek-V2} using prompt \textbf{\texttt{P2}}. Challenging candidates are manually selected as the final task queries. Both sets consist of 1,000 curated samples, equally split into 500 CPQs and 500 true presupposition questions (TPQs). All counterfactual candidates in CPQs of CP-Bench are chosen from objects commonly associated with similar semantics or scenes, such as "\texttt{railroad}" in train-related contexts. This increases the benchmark's complexity, as prior research has shown that LVLMs often suffer from inherent statistical biases in their pre-training or fine-tuning datasets~\citep{li2023evaluating, yu2024hallucidoctor}. The samples in CP-Bench can be classified into four categories in daily scenarios: \textit{item}, \textit{knowledge}, \textit{scene}, and \textit{activity}. More statistical details of CP-Bench can be viewed in Figure \ref{fig_cpbench}.

\vspace{1pt}

\noindent \textbf{Evaluation.} Given the open-ended responses of LVLMs, \texttt{GPT-4o}~\citep{achiam2023gpt} is introduced to convert responses into a binary classification task, assessing whether the models correctly recognize the correctness presupposition and output factual responses. CPQ labeled as “\texttt{positive}” class, while TPQs are labeled as “\texttt{negative}” samples. For CPQs, \texttt{GPT-4o} evaluates whether the models accurately identify the presence of the objects and generate corresponding responses. For TPQs, \texttt{GPT-4o} evaluates whether the models can discriminate and the reveal counterfactual information implicit in presuppositions. The primary evaluation metrics are the \textit{F1-score}, \textit{Recall}, \textit{Accuracy}, and \textit{Precision}. The prompt \textbf{\texttt{P3}} used for CP-Bench evaluation is provided in \textit{Appendix}.

\section{Experiment}

\subsection{Implement Setup}

\noindent \textbf{Experiment Baselines:} We post-trained LLaVA series with the proposed \textit{Antidote}, including LLaVA-1.5-Vincuna-7B/13B~\citep{liu2024visual}, and LLaVA-Next-Mistral-7B~\citep{liu2024llavanext}. All models above have been fully tuned on their collected visual instruction data before post-training. In practical implementation, we adopt LoRA~\citep{hu2021lora} for training efficiency. The LoRA's dimension (rank) $r$ is 64, $\alpha$ is 128, and the scale parameter $\beta$ in direct preference optimization is 0.1. More hyper-parameter setting can be viewed in \textit{Appendix}.

\vspace{2pt}

\noindent \textbf{Evaluation Benchmarks.} Besides CP-Bench, we assess the effectiveness of \textit{Antidote} using three popular hallucination benchmarks and four general benchmarks. POPE~\citep{li2023evaluating} is used for evaluating object existence, while CHAIR~\citep{rohrbach2018object} and SHR~\citep{zhao2023beyond} are served to evaluate the hallucinations in image descriptions. Compared to CHAIR, which focuses on evaluating object-related hallucinations in responses, SHR focuses on sentence-level hallucinations with the introduction of LLMs. To further validate the catastrophic forgetting issue, we verify the general capability (such as visual reasoning, perception, and cross-domain generalization) of models trained with \textit{Antidote}, including Science-QA~\citep{saikh2022scienceqa}, MMBench~\citep{liu2023mmbench}, MMVet~\citep{yu2023mm}, and LLaVA-Wild~\citep{liu2024visual}.

\vspace{2pt}

\noindent \textbf{Data Composition.} The training set of \textit{Antidote} is built by the data synthetic pipeline introduced in Section \ref{sect_synthetic}. Initially, we generated \textit{14k} triplets of $\textless \mathcal{C}_{img}, \mathcal{O}_{pre}, \mathcal{O}_{hallu} \textgreater$ and filtered out approximately \textit{4k} triplets with \textit{Factual Accessor}. Then, we generated the queries of CPQs, TPQs, object existence, and image descriptions of each remaining triplet. For each baseline LVLM, we applied response filtering after their inference and self-correction, discarding around 15\% of the total samples. Finally, we sample 5,000 CPQs, 5,000 TPQs, 2,000 questions of object existence, and 8,000 image description queries, \ie, a total of \textit{20k} samples, for post-training. The discussion of data proportion settings can be viewed in \textit{Appendix}.

\subsection{Main Results}

\noindent \textbf{How does \textit{Antidote} improve LVLMs' ability to \textit{discriminate presuppositions}?} In Table \ref{table_fpvqa_merged}, we compare the performance of closed-sourced models, open-sourced models, and baseline models post-trained with \textit{Antidote} on CP-Bench. The results reveal that the closed-source LVLMs significantly outperform the open-source models in distinguishing CPQ and outputting factual responses. The optimal performance is achieved by Claude-3.5-Sonnet and GPT-4o, which recall nearly 90\% of CPQs on the \textit{test} set. For open-source models, LLaVA-Next-Vicuna-13B stands out, achieving an F1-score of 56.0\% and a recall of 41.6\%. Notably, \textit{Antidote} brings substantial improvements. For example, it 
boosts LLaVA-1.5 7B and 13B models' F1-scores from 5.7 and 17.3 to 78.4 and 83.5, respectively. As illustrated in Figure \ref{fig_fpq}, we can observe that these models after \textit{Antidote} can produce factual responses such as “\textit{I cannot answer ... as there is no ...”}. These results highlight \textit{Antidote}'s efficacy in enhancing model accuracy in recognizing presuppositions, pushing open-source models closer to closed-source counterparts on this challenging task.

\begin{table*}[t]
\scriptsize
\renewcommand\arraystretch{1.14}
\setlength{\tabcolsep}{1.0mm}
\centering
\begin{minipage}{0.59\textwidth}
\centering
\begin{tabular}{l|c|ccc|c}
\toprule[1.2pt]
\multicolumn{1}{c|}{\multirow{2}{*}{\textbf{Method}}} & \multicolumn{4}{c|}{\textbf{CP-Bench-Test}} & \textbf{CP-Bench-Dev} \\ \cline{2-6} 
\multicolumn{1}{c|}{} & \textbf{F1-Score} & \textbf{Accuracy} & \textbf{Precision} & \multicolumn{1}{c|}{\textbf{Recall}} & \textbf{F1-Score} \\ \midrule
\multicolumn{6}{l}{\textit{\textbf{Closed-sourced (API)}}} \\
\rowcolor[HTML]{eff6fc}
Claude-3-5-Sonnet~\citep{claude3_5}       & 86.0 & 85.3 & 82.4 & 90.0  & 94.3 \\
GPT-4o-0806~\citep{gpt_4o}                & 85.5 & 85.0 & 82.5 & 88.8  & 84.2 \\
GPT-4o-mini-0718~\citep{gpt_4o}           & 77.6 & 75.9 & 72.6 & 83.3  & 81.8 \\
GLM-4v~\citep{glm2024chatglm}             & 79.7 & 80.3 & 82.1 & 79.7  & 88.2 \\
GPT-4v-0409~\citep{achiam2023gpt}         & 71.6 & 75.8 & 86.6 & 61.0  & 86.0 \\
InternVL-2-Pro~\citep{chen2024internvl}   & 64.2 & 70.5 & 81.5 & 53.0  & 60.3 \\ \midrule
\multicolumn{6}{l}{\textit{\textbf{Open-sourced}}} \\
\rowcolor[HTML]{eff6fc}
LLaVA-Next-Vicuna-13B~\citep{liu2024llavanext} & 56.0 & 67.3 & 85.5 & 41.6  & 65.1 \\
MiniCPM-V2.5-8B~\citep{yao2024minicpm}    & 55.2 & 66.1 & 81.3 & 41.8  & 37.0 \\
Qwen2-VL-72B~\citep{wang2024qwen2}           & 48.8 & 65.0 & 90.7 & 33.3  & 42.6    \\
LLaVA-Next-Vicuna-7B~\citep{liu2024llavanext}  & 47.0 & 63.6 & 86.1 & 32.3  & 48.7 \\
InternVL2-26B~\citep{chen2024internvl}    & 38.5 & 59.6 & 80.8 & 25.3  & 41.2 \\
InstructBLIP-7B~\citep{instructblip}      & 35.6 & 47.4 & 45.9 & 29.1  & 17.8 \\
InternVL2-8B~\citep{chen2024internvl}     & 34.3 & 58.4 & 81.8 & 21.7  & 47.1 \\
Cogvlm2-19B~\citep{hong2024cogvlm2}       & 34.2 & 57.4 & 75.3 & 22.1  & 42.8 \\
Phi-3-Vision~\citep{abdin2024phi}           & 26.1 & 54.0 & 66.4 & 16.3  & 38.8    \\
Qwen2-VL-7B~\citep{wang2024qwen2}             & 22.7 & 54.9 & 80.3 & 13.2  & 33.7    \\ \midrule
\multicolumn{6}{l}{\textit{\textbf{Baseline + Post-training / Decoding Strategies}}} \\
\textbf{LLaVA-v1.5-7B}~\citep{liu2024improved}     & 5.7  & 50.5 & 60.0 & 3.0   & 12.4 \\
+ HA-DPO~\citep{zhao2023beyond}             & 4.7  & 50.6 & 66.7 & 2.4   & 13.1 \\
+ VCD~\citep{zhu2024self}                  & 6.1 & 50.7 & 64.0 & 3.2  & 12.0 \\
+ SeVa~\citep{zhu2024self}                  & 24.1 & 55.1 & 78.0 & 14.3  & 25.4 \\
\rowcolor[HTML]{eff6fc}
+ \textbf{\textit{Antidote}} (ours)                         & \textbf{78.4} \textbf{\textcolor{ForestGreen}{(+72.7)}} & \textbf{84.5} \textbf{\textcolor{ForestGreen}{(+34.0)}} & \textbf{73.1} \textbf{\textcolor{ForestGreen}{(+13.1)}} & \textbf{73.1} \textbf{\textcolor{ForestGreen}{(+70.1)}} & \textbf{82.9 \textcolor{ForestGreen}{(+70.5)}} \\
\midrule
\textbf{LLaVA-v1.5-13B}~\citep{liu2024improved}    & 17.3 & 53.8 & 82.8 & 9.6   & 12.0 \\
\rowcolor[HTML]{eff6fc}
+ \textbf{\textit{Antidote}} (ours)                       & \textbf{83.5} \textbf{\textcolor{ForestGreen}{(+66.2)}} & \textbf{84.5} \textbf{\textcolor{ForestGreen}{(+31.3)}} & \textbf{89.5} \textbf{\textcolor{ForestGreen}{(+6.7)}} & \textbf{78.3} \textbf{\textcolor{ForestGreen}{(+69.7)}} & \textbf{88.0 \textcolor{ForestGreen}{(+66.0)}} \\
\midrule
\textbf{LLaVA-Next-Mistral-7B}~\citep{liu2024llavanext} & 26.7 & 54.8 & 70.7 & 16.5  & 43.6 \\
\rowcolor[HTML]{eff6fc}
+ \textbf{\textit{Antidote}} (ours)                        & \textbf{76.8} \textbf{\textcolor{ForestGreen}{(+50.1)}} & \textbf{77.5} \textbf{\textcolor{ForestGreen}{(+22.7)}} & \textbf{79.4} \textbf{\textcolor{ForestGreen}{(+8.7)}}  & \textbf{74.3} \textbf{\textcolor{ForestGreen}{(+58.8)}} & \textbf{84.4 \textcolor{ForestGreen}{(+40.8)}} \\
\bottomrule[1.2pt]
\end{tabular}
\vspace{-4pt}
\caption{\textbf{Performance Comparison on CP-Bench (\textit{test} and \textit{dev}).} The detailed evaluation results of the \textit{dev} set are provided in \textit{Appendix}.}
\vspace{-10pt}
\label{table_fpvqa_merged}
\end{minipage}%
\hfill
\begin{minipage}{0.36\textwidth}
\centering
\vspace{5pt}
\begin{minipage}{\textwidth}
\centering
\small
\renewcommand\arraystretch{1.05}
\setlength{\tabcolsep}{2.8mm}
\scriptsize
\begin{tabular}{l|cc}
\toprule[1.2pt]
{\textbf{Method}}                & \textbf{Accuracy ($\%$)} & \textbf{F1-Score ($\%$)} \\ \midrule
\textbf{LLaVA-1.5-7B}                     & 85.18             & 86.07             \\
+ VDD~\citep{liu2024visual}                  & 86.47             & 85.13             \\
+ HACL~\citep{jiang2024hallucination}               & 86.66             & 86.20             \\
+ Volcano~\citep{lee2023volcano}             & 86.96             & 86.67             \\
+ HA-DPO~\citep{zhao2023beyond}              & 86.63             & 86.87             \\
+ SeVa~\citep{zhu2024self}                & 86.69             & 86.66             \\
\rowcolor[HTML]{eff6fc}
+ \textbf{\textit{Antidote}} (ours)            & \textbf{88.09 \textcolor{ForestGreen}{(+3.75)}}    & \textbf{87.89 \textcolor{ForestGreen}{(+1.82)}}    \\ \midrule
\textbf{LLaVA-1.5-13B}                   & 84.15             & 85.67             \\
+ Volcano~\citep{lee2023volcano}             & 87.02             & 87.17             \\
\rowcolor[HTML]{eff6fc}
+ \textbf{\textit{Antidote}} (ours)            & \textbf{88.93 \textcolor{ForestGreen}{(+4.78)}}    & \textbf{88.99 \textcolor{ForestGreen}{(+4.32)}}    \\ \bottomrule[1.2pt]
\end{tabular}
\vspace{-4pt}
\caption{\textbf{Hallucination evaluation (POPE) on object existence.} The performance is the \textit{average} of the results across the three subsets. }
\vspace{-4pt}
\label{table_pope}
\end{minipage}

\vspace{12pt} 
\begin{minipage}{\textwidth}
\centering
\small
\renewcommand\arraystretch{1.06}
\setlength{\tabcolsep}{0.15mm}
\scriptsize
\begin{tabular}{l|ccc}
\toprule[1.2pt]
\textbf{Method}                & \textbf{CHAIR\_s $\downarrow$} & \textbf{CHAIR\_i $\downarrow$} & \textbf{SHR $\downarrow$}  \\ \midrule
\textbf{LLaVA-1.5-7B}          & 19.4              & 6.1               & 36.7          \\
+ SeVa~\citep{zhu2024self}                           & 18.4              & 5.7               & 34.9          \\
+ VCD~\citep{liu2024visual}                            & 17.9              & 5.8               & 34.2          \\
+ OPERA~\citep{huang2024opera}                          & 15.6              & 5.7               & 34.1          \\
+ HA-DPO~\citep{zhao2023beyond}                        & 18.0              & 5.9               & 34.0          \\
+ SID~\citep{huo2024self}                            & 15.1              & 5.4               & 33.1          \\
\rowcolor[HTML]{eff6fc}
+ \textbf{\textit{Antidote}}              & \textbf{9.4 \textcolor{ForestGreen}{(-10.0)}}      & \textbf{3.3 \textcolor{ForestGreen}{(-2.8)}}      & \textbf{18.1 \textcolor{ForestGreen}{(-18.6)}} \\ \midrule
\textbf{LLaVA-1.5-13B}         & 30.0              & 5.5               & 37.2          \\
\rowcolor[HTML]{eff6fc}
+ \textbf{\textit{Antidote}} (ours)             & \textbf{12.6 \textcolor{ForestGreen}{(-17.4)}}     & \textbf{4.3 \textcolor{ForestGreen}{(-1.2)}}      & \textbf{21.3 \textcolor{ForestGreen}{(-15.9)}} \\ \midrule
\textbf{LLaVA-Next-Mistral-7B} & 13.0              & 4.6               & 28.4          \\
\rowcolor[HTML]{eff6fc}
+ \textbf{\textit{Antidote}} (ours)             & \textbf{10.7 \textcolor{ForestGreen}{(-2.3)}}     & \textbf{3.5 \textcolor{ForestGreen}{(-1.1)}}      & \textbf{19.7 \textcolor{ForestGreen}{(-8.7)}} \\ \bottomrule[1.2pt]
\end{tabular}
\vspace{-4pt}
\caption{\textbf{Hallucination evaluation on image description, CHAIR and SHR.} \textit{Max new tokens} is set as 64 for each model. Smaller values correspond to fewer hallucinations. }
\vspace{-4pt}
\label{table_caption}
\end{minipage}
\end{minipage}
\end{table*}

\vspace{1pt}

\noindent \textbf{Does \textit{model size} affect the ability to discriminate the correctness of presuppositions?} (1) \textit{With identical architectures and training data, larger model sizes enhance the judgment of presupposition correctness.} For instance, as Qwen2-VL ~\citep{wang2024qwen2} scales from 7B to 72B parameters, the recall increases from 13.2\% to 33.3\%, with a similar trend observed in the InternVL2 series. (2) \textit{Across different models, however, model size is not a decisive factor.} Notably, MiniCPM-V2.5~\citep{yao2024minicpm}, with only 8B parameters, achieves a recall that is 8.5\% higher than Qwen2-VL-72B, demonstrating superior performance in recognizing CPQs. Moreover, InternVL2~\citep{chen2024internvl} and Qwen2-VL, which surpass closed-source models in general performance, do not perform well on the VPF-Bench. Both models have utilized large-scale instruction fine-tuning datasets to enhance visual capabilities. We believe that their performance on the VPF-Bench is \textit{strongly correlated with over-learning of instruction tuning}.

\vspace{2pt}

\noindent \textbf{How do LVLMs with \textit{Antidote} perform on typical hallucination benchmarks?} Here, we compare various types of mitigation approaches, including contrastive decoding (\eg, VCD~\citep{liu2024visual} and VDD~\citep{zhang2024debiasing}), auxiliary learning (\eg, HACL~\citep{jiang2024hallucination}), and post-training (\eg, Volcano~\citep{lee2023volcano} and SeVA~\citep{zhu2024self}). \textbf{1) For object existence}, we assess POPE, where the results (Table \ref{table_pope}) are averaged across three evaluation sets: the \textit{random}, \textit{popular}, and \textit{adversarial} sets (the results for each set can be found in \textit{Appendix}. On LLaVA 1.5-7B, we improved its original F1-score \textbf{from 86.07 to 87.89 (+1.82\%)}, with an even greater improvement on its 13B version, \textbf{from 85.67 to 88.99 (+3.32\%)}. Notably, we observed significant improvements on the \textit{adversarial} subset, where objects are first ranked based on co-occurrence frequencies, and the top-k frequent objects are sampled. On the original 7B and 13B versions, \textbf{\textit{Antidote} improves by 2.58\% and 4.12\%}, respectively. This demonstrates that \textit{Antidote} can effectively mitigate the statistical biases inherent in LVLMs, which substantially contribute to object hallucination issues~\citep{li2023evaluating}. \textbf{2) For image description}, we first evaluated on CHAIR (Table \ref{table_caption}), which quantifies the hallucination by calculating the ratio of objects mentioned in the description that are not present in the ground-truth. On the LLaVA-1.5 series, we observed a substantial reduction in hallucinations, \textbf{decreasing its hallucination rates by over 50\%}. For the 7B version, we reduced CHAIR\_s from the prior best score of \textbf{15.1 to 9.4}. We also tested on LLaVA-Next-Mistral-7B, further improving its CHAIR\_s and CHAIR\_i scores to \textbf{10.7 and 3.5}, respectively. Additionally, we evaluated SHR~\citep{zhao2023beyond}, an advanced benchmark that uses detailed object-level descriptions from the Visual-Genome dataset as factual information and relies on GPT-4 to judge hallucinations in descriptions. Similarly, \textit{Antidote} significantly reduces hallucinations in comparison to baseline models on this metric as well.

\begin{table*}[!t]
\centering
\begin{minipage}{0.6\textwidth}
\scriptsize
\centering
\renewcommand\arraystretch{1.0}
\setlength{\tabcolsep}{1.1mm}
\begin{tabular}{l|ccc|cccc}
\toprule[1.2pt]
\textbf{Method} & \textbf{POPE}  & \textbf{CP-Bench} & \textbf{SHR} ($\downarrow$) & \textbf{Science-QA} & \textbf{MMBench} & \textbf{MMVet} & \textbf{LLaVA$^\text{W}$} \\ \midrule
LLaVA-1.5-7B    & 85.92          & 12.4                & 36.7           & 66.8                & 64.3             & 30.5           & \textbf{65.4}       \\
\rowcolor[HTML]{eff6fc}
+ \textbf{\textit{Antidote}} (ours)   & \textbf{87.89} & \textbf{81.2}       & \textbf{18.1}  & \textbf{69.6}       & \textbf{65.4}    & \textbf{31.4}  & {64.0}          \\ \midrule
LLaVA-1.5-13B   & 85.67          & 12.0                & 37.2           & 71.6                & 67.7             & 35.4           & \textbf{70.7}       \\
\rowcolor[HTML]{eff6fc}
+ \textbf{\textit{Antidote}} (ours)   & \textbf{88.99} & \textbf{88.0}       & \textbf{21.3}  & \textbf{74.2}       & \textbf{69.5}    & \textbf{35.5}  & {70.2}          \\ \bottomrule[1.2pt]
\end{tabular}
\vspace{-4pt}
\caption{\textbf{The evaluation on the benchmark of LVLMs' general capabilities.}}
\label{tab_cata}
\vspace{10pt} 
\centering
\captionsetup{type=figure}
\includegraphics[width=1\textwidth]{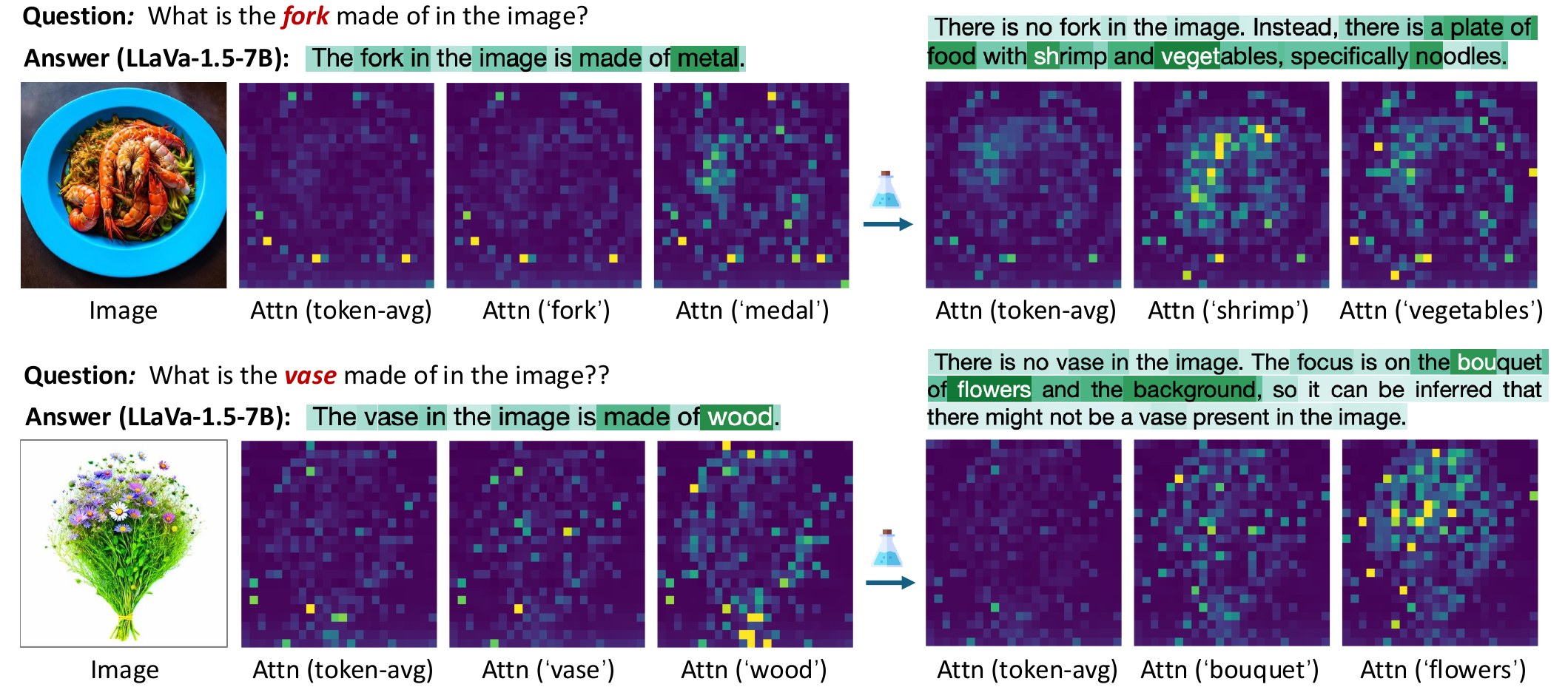}
\vspace{-12pt}
\caption{\textbf{Attention visualization between the text tokens and vision tokens.} The intensity of each text token’s background indicates the attention weight magnitude of image tokens, with darker highlights representing higher attention. The attention values above the 0.995th quantile are shown with the highest color intensity (such as \texttt{shrimp} and \texttt{vegetables}).}
\vspace{-6pt}
\label{fig_attn}
\end{minipage}
\hfill
\begin{minipage}{0.37\textwidth}
\scriptsize
\centering
\renewcommand\arraystretch{1.05}
\setlength{\tabcolsep}{1.5mm}
\vspace{5pt}
\begin{tabular}{cc|cccc}
\toprule[1.2pt]
$r$     & $a$     & \textbf{CP-Bench}    & \textbf{POPE}     & \textbf{SHR ($\downarrow$)}     & \textbf{MMBench}    \\ \midrule
\multicolumn{2}{c|}{\textit{Baseline}} & 12.4      & 86.07    & 36.7    & 64.3       \\
\multicolumn{2}{c|}{\textit{\textbf{SFT}}} & 67.8      & 85.14    & 25.9    & 59.6       \\ \midrule

32         & 64         & 12.7      & 86.68    & 28.5    & 64.1       \\
\rowcolor[HTML]{eff6fc}
\textbf{64}         & \textbf{128}        & \textbf{82.9}      & \textbf{87.89}    & \textbf{18.1}    & \textbf{65.4}       \\
128        & 256        & 87.8      & 85.67    & 23.9    & 47.3       \\ \midrule
256        & 512        & \multicolumn{4}{c}{{\textit{Model Collapse}}} \\ \bottomrule[1.2pt]
\end{tabular}
\vspace{-4pt}
\caption{The effect of \textbf{hyper-parameter in LoRA} during the post-training and the \textbf{SFT alternative}. The baseline LVLM is LLaVA-1.5-7B. F1-score is adopted in CP-Bench and POPE.}
\label{lora_sft}
\vspace{6pt} 
\captionsetup{type=figure}
\vspace{6pt}
\includegraphics[width=1.0\textwidth]{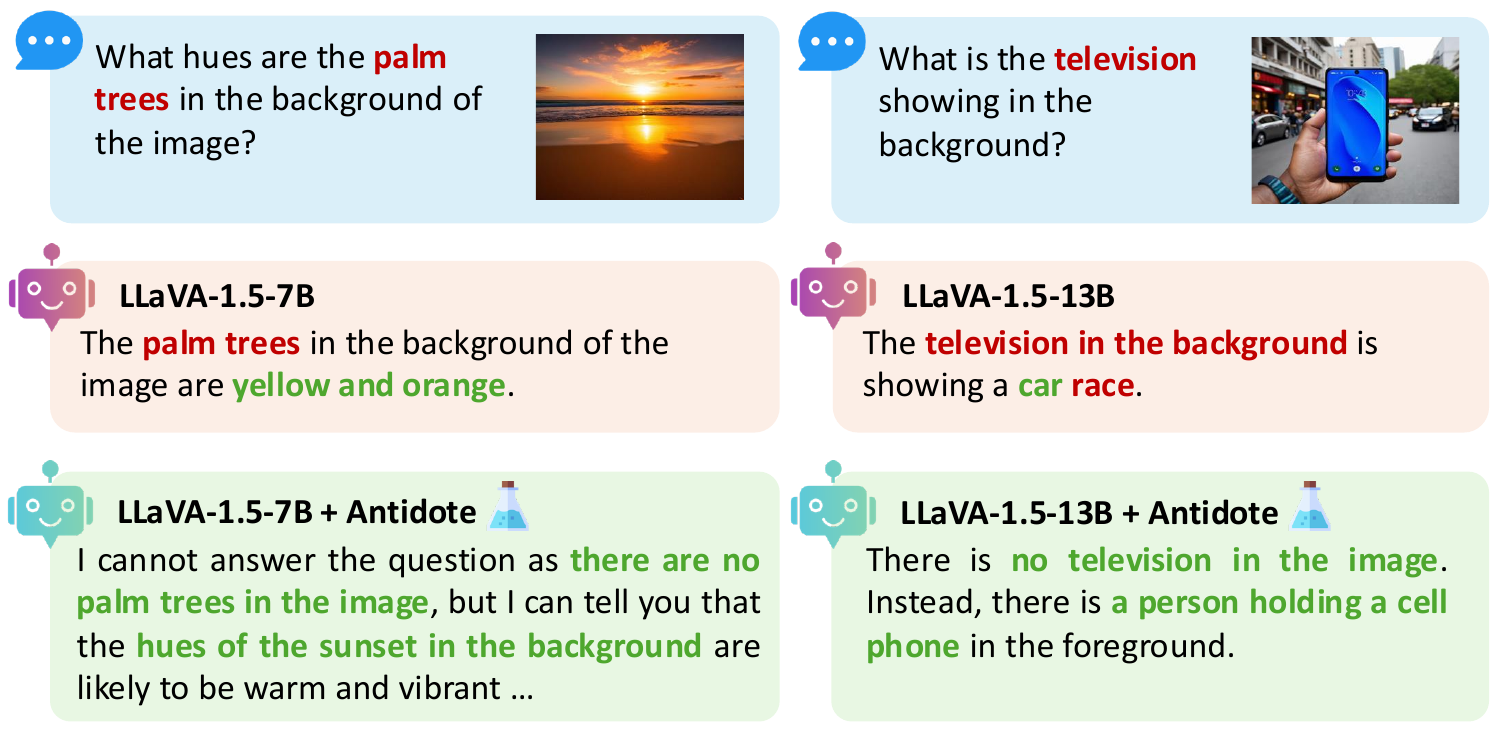}
\vspace{-10pt}
\caption{\textbf{Comparison of model responses before and after \textit{Antidote} post-training.} The cases are selected from the proposed CP-Bench benchmark.}
\label{fig_fpq}
\end{minipage}
\vspace{-4pt}
\end{table*}

\subsection{Analysis}

\noindent \textbf{Catastrophic Forgetting.} Since \textit{Antidote} is a post-training method that fine-tunes the baseline models' parameters, we evaluated whether \textit{Antidote} causes catastrophic forgetting by assessing the general capability of the post-trained LLaVA-1.5 series. From Table \ref{tab_cata}, it is evident that performance on these benchmarks did not significantly degrade and even improved on some benchmarks, such as a 2.8\% and 2.6\% increase on Science-QA \cite{lu2022learn}. This suggests \textbf{suppressing object perception hallucinations and enhancing CPQ discrimination can generalize to improvements in overall capabilities}. There was a slight decrease in performance on LLaVA-Wild \cite{liu2024visual}, where we observed that the post-trained version was “cautious” when answering uncertain/challenging questions compared to the baseline model, which is not preferred by its \texttt{GPT-4} evaluator.

\vspace{2pt}

\noindent \textbf{Attention Visualization.} We empirically investigate how attention from visual tokens contributes to important object-related text tokens before and after applying the proposed \textit{Antidote}. We visualize some representative instances during training in Figure \ref{fig_attn}. For instance, when asked a CPQ: “\textit{What is the fork made of in the image?}”, we observe that the original LLaVA-1.5, while outputting “\texttt{fork}”, does not significantly focus on visual tokens, and incorrectly attends to visual token information when outputting “\texttt{metal}”. However, after training with \textit{Antidote}, the model's attention to visual tokens becomes more accurate, focusing on the exact areas of the image corresponding to object-related text tokens, such as “\texttt{shrimp}” and “\texttt{flowers}”.

\vspace{2pt}

\noindent \textbf{Compare with SFT.} A straightforward alternative to \textit{Antidote}'s preference optimization is direct supervised fine-tuning (SFT) using the self-corrected responses in \textit{Antidote}. As shown in Table \ref{lora_sft}, while SFT shows effectiveness in addressing model hallucinations for CPQs and image descriptions, \textbf{it significantly underperforms compared to \textit{Antidote}}, particularly on POPE and MMBench, and suffers from catastrophic forgetting to some extent. Unlike SFT, which merely increases the probability of self-corrected responses, \textit{Antidote}'s preference alignment can be viewed as a form of contrastive learning (more discussions can be viewed in \textit{Appendix}), where the model is trained to distinguish between self-corrected and hallucinatory responses. It exploits the preference information by increasing the model's probability of self-corrected responses relative to hallucinatory ones, guiding the model to suppress hallucinations while reducing over-fitting to preference samples.

\vspace{2pt}

\noindent \textbf{LoRA Fine-tuning.} We evaluate the LoRA setting for \textit{Antidote}. In parameter-efficient learning, this parameter determines the extent to which the model’s knowledge can be altered during post-training. As presented in Table \ref{lora_sft}, a relatively higher rank $r$ signifies greater flexibility in adjusting the model’s knowledge. However, a larger $r$ can lead to catastrophic forgetting (when $r$=128) and even cause over-optimization of \textit{Antidote}, resulting in model collapse (when $r$=256). In conclusion, we set the rank $r$ to 64 and the scaling factor $a$ to 128 (2$\times r$ as default) in experiments.

\section{Conclusion}

This paper addresses the hallucinations in Large Vision-Language Models (LVLMs), focusing both on Counterfactual Presupposition Questions (CPQs) and object perception. The primary contribution of this work is Antidote, a synthetic data-driven post-training framework that significantly mitigates hallucinations by enhancing LVLMs' capability to discern counterfactual presuppositions and objects. To complement this, we introduce CP-Bench, a novel benchmark tailored to evaluate LVLMs' performance on CPQs. Extensive experiments demonstrate that Antidote not only significantly improves performance in recognizing counterfactual presuppositions but also enhances performance across a range of hallucination-related benchmarks, including POPE, CHAIR, and SHR, all without inducing catastrophic forgetting. These results underscore the promise of \textit{Antidote} as a robust solution for enhancing the reliability of LVLMs in diverse vision-language tasks.

{
    \small
    \bibliographystyle{ieeenat_fullname}
    \bibliography{main}

\begin{thebibliography}{51}
\providecommand{\natexlab}[1]{#1}
\providecommand{\url}[1]{\texttt{#1}}
\expandafter\ifx\csname urlstyle\endcsname\relax
  \providecommand{\doi}[1]{doi: #1}\else
  \providecommand{\doi}{doi: \begingroup \urlstyle{rm}\Url}\fi

\bibitem[Abdin et~al.(2024)Abdin, Jacobs, Awan, Aneja, Awadallah, Awadalla, Bach, Bahree, Bakhtiari, Behl, et~al.]{abdin2024phi}
Marah Abdin, Sam~Ade Jacobs, Ammar~Ahmad Awan, Jyoti Aneja, Ahmed Awadallah, Hany Awadalla, Nguyen Bach, Amit Bahree, Arash Bakhtiari, Harkirat Behl, et~al.
\newblock Phi-3 technical report: A highly capable language model locally on your phone.
\newblock \emph{arXiv preprint arXiv:2404.14219}, 2024.

\bibitem[Achiam et~al.(2023)Achiam, Adler, Agarwal, Ahmad, Akkaya, Aleman, Almeida, Altenschmidt, Altman, Anadkat, et~al.]{achiam2023gpt}
Josh Achiam, Steven Adler, Sandhini Agarwal, Lama Ahmad, Ilge Akkaya, Florencia~Leoni Aleman, Diogo Almeida, Janko Altenschmidt, Sam Altman, Shyamal Anadkat, et~al.
\newblock Gpt-4 technical report.
\newblock \emph{arXiv preprint arXiv:2303.08774}, 2023.

\bibitem[Anthropic(2024)]{claude3_5}
Anthropic.
\newblock Claude 3.5 sonnet model card addendum.
\newblock \url{https://www-cdn.anthropic.com/fed9cc193a14b84131812372d8d5857f8f304c52/Model_Card_Claude_3_Addendum.pdf}, 2024.

\bibitem[Chen et~al.(2024{\natexlab{a}})Chen, Xiao, Zhang, Luo, Lian, and Liu]{chen2024bge}
Jianlv Chen, Shitao Xiao, Peitian Zhang, Kun Luo, Defu Lian, and Zheng Liu.
\newblock Bge m3-embedding: Multi-lingual, multi-functionality, multi-granularity text embeddings through self-knowledge distillation.
\newblock \emph{arXiv preprint arXiv:2402.03216}, 2024{\natexlab{a}}.

\bibitem[Chen et~al.(2024{\natexlab{b}})Chen, Wu, Wang, Su, Chen, Xing, Zhong, Zhang, Zhu, Lu, et~al.]{chen2024internvl}
Zhe Chen, Jiannan Wu, Wenhai Wang, Weijie Su, Guo Chen, Sen Xing, Muyan Zhong, Qinglong Zhang, Xizhou Zhu, Lewei Lu, et~al.
\newblock Internvl: Scaling up vision foundation models and aligning for generic visual-linguistic tasks.
\newblock In \emph{Proceedings of the IEEE/CVF Conference on Computer Vision and Pattern Recognition}, pages 24185--24198, 2024{\natexlab{b}}.

\bibitem[Dai et~al.(2023)Dai, Li, Li, Tiong, Zhao, Wang, Li, Fung, and Hoi]{instructblip}
Wenliang Dai, Junnan Li, Dongxu Li, Anthony Meng~Huat Tiong, Junqi Zhao, Weisheng Wang, Boyang Li, Pascale Fung, and Steven Hoi.
\newblock Instructblip: Towards general-purpose vision-language models with instruction tuning, 2023.

\bibitem[Esser et~al.(2024)Esser, Kulal, Blattmann, Entezari, M{\"u}ller, Saini, Levi, Lorenz, Sauer, Boesel, et~al.]{esser2024scaling}
Patrick Esser, Sumith Kulal, Andreas Blattmann, Rahim Entezari, Jonas M{\"u}ller, Harry Saini, Yam Levi, Dominik Lorenz, Axel Sauer, Frederic Boesel, et~al.
\newblock Scaling rectified flow transformers for high-resolution image synthesis.
\newblock In \emph{Forty-first International Conference on Machine Learning}, 2024.

\bibitem[GLM-Team et~al.(2024)GLM-Team, Zeng, Xu, Wang, Zhang, Yin, Rojas, Feng, Zhao, Lai, et~al.]{glm2024chatglm}
GLM-Team, Aohan Zeng, Bin Xu, Bowen Wang, Chenhui Zhang, Da Yin, Diego Rojas, Guanyu Feng, Hanlin Zhao, Hanyu Lai, et~al.
\newblock Chatglm: A family of large language models from glm-130b to glm-4 all tools.
\newblock \emph{arXiv preprint arXiv:2406.12793}, 2024.

\bibitem[Hong et~al.(2024)Hong, Wang, Ding, Yu, Lv, Wang, Cheng, Huang, Ji, Xue, et~al.]{hong2024cogvlm2}
Wenyi Hong, Weihan Wang, Ming Ding, Wenmeng Yu, Qingsong Lv, Yan Wang, Yean Cheng, Shiyu Huang, Junhui Ji, Zhao Xue, et~al.
\newblock Cogvlm2: Visual language models for image and video understanding.
\newblock \emph{arXiv preprint arXiv:2408.16500}, 2024.

\bibitem[Hu et~al.(2021)Hu, Shen, Wallis, Allen-Zhu, Li, Wang, Wang, and Chen]{hu2021lora}
Edward~J Hu, Yelong Shen, Phillip Wallis, Zeyuan Allen-Zhu, Yuanzhi Li, Shean Wang, Lu Wang, and Weizhu Chen.
\newblock Lora: Low-rank adaptation of large language models.
\newblock \emph{arXiv preprint arXiv:2106.09685}, 2021.

\bibitem[Huang et~al.(2024)Huang, Dong, Zhang, Wang, He, Wang, Lin, Zhang, and Yu]{huang2024opera}
Qidong Huang, Xiaoyi Dong, Pan Zhang, Bin Wang, Conghui He, Jiaqi Wang, Dahua Lin, Weiming Zhang, and Nenghai Yu.
\newblock Opera: Alleviating hallucination in multi-modal large language models via over-trust penalty and retrospection-allocation.
\newblock In \emph{Proceedings of the IEEE/CVF Conference on Computer Vision and Pattern Recognition}, pages 13418--13427, 2024.

\bibitem[Huo et~al.(2024)Huo, Xu, Zhang, Wang, Chen, and Zhao]{huo2024self}
Fushuo Huo, Wenchao Xu, Zhong Zhang, Haozhao Wang, Zhicheng Chen, and Peilin Zhao.
\newblock Self-introspective decoding: Alleviating hallucinations for large vision-language models.
\newblock \emph{arXiv preprint arXiv:2408.02032}, 2024.

\bibitem[Jafari et~al.(2021)Jafari, Maurya, Nagarkar, Islam, and Crushev]{jafari2021survey}
Omid Jafari, Preeti Maurya, Parth Nagarkar, Khandker~Mushfiqul Islam, and Chidambaram Crushev.
\newblock A survey on locality sensitive hashing algorithms and their applications.
\newblock \emph{arXiv preprint arXiv:2102.08942}, 2021.

\bibitem[Ji et~al.(2023)Ji, Yu, Xu, Lee, Ishii, and Fung]{ji2023towards}
Ziwei Ji, Tiezheng Yu, Yan Xu, Nayeon Lee, Etsuko Ishii, and Pascale Fung.
\newblock Towards mitigating llm hallucination via self reflection.
\newblock In \emph{Findings of the Association for Computational Linguistics: EMNLP 2023}, pages 1827--1843, 2023.

\bibitem[Jiang et~al.(2024)Jiang, Xu, Dong, Chen, Ye, Yan, Ye, Zhang, Huang, and Zhang]{jiang2024hallucination}
Chaoya Jiang, Haiyang Xu, Mengfan Dong, Jiaxing Chen, Wei Ye, Ming Yan, Qinghao Ye, Ji Zhang, Fei Huang, and Shikun Zhang.
\newblock Hallucination augmented contrastive learning for multimodal large language model.
\newblock In \emph{Proceedings of the IEEE/CVF Conference on Computer Vision and Pattern Recognition}, pages 27036--27046, 2024.

\bibitem[Lee et~al.(2023)Lee, Park, Jo, and Seo]{lee2023volcano}
Seongyun Lee, Sue~Hyun Park, Yongrae Jo, and Minjoon Seo.
\newblock Volcano: mitigating multimodal hallucination through self-feedback guided revision.
\newblock \emph{arXiv preprint arXiv:2311.07362}, 2023.

\bibitem[Leng et~al.(2024)Leng, Zhang, Chen, Li, Lu, Miao, and Bing]{leng2024mitigating}
Sicong Leng, Hang Zhang, Guanzheng Chen, Xin Li, Shijian Lu, Chunyan Miao, and Lidong Bing.
\newblock Mitigating object hallucinations in large vision-language models through visual contrastive decoding.
\newblock In \emph{Proceedings of the IEEE/CVF Conference on Computer Vision and Pattern Recognition}, pages 13872--13882, 2024.

\bibitem[Li et~al.(2023)Li, Du, Zhou, Wang, Zhao, and Wen]{li2023evaluating}
Yifan Li, Yifan Du, Kun Zhou, Jinpeng Wang, Wayne~Xin Zhao, and Ji-Rong Wen.
\newblock Evaluating object hallucination in large vision-language models.
\newblock \emph{arXiv preprint arXiv:2305.10355}, 2023.

\bibitem[Liu et~al.(2024{\natexlab{a}})Liu, Feng, Wang, Wang, Liu, Zhao, Dengr, Ruan, Dai, Guo, et~al.]{liu2024deepseek}
Aixin Liu, Bei Feng, Bin Wang, Bingxuan Wang, Bo Liu, Chenggang Zhao, Chengqi Dengr, Chong Ruan, Damai Dai, Daya Guo, et~al.
\newblock Deepseek-v2: A strong, economical, and efficient mixture-of-experts language model.
\newblock \emph{arXiv preprint arXiv:2405.04434}, 2024{\natexlab{a}}.

\bibitem[Liu et~al.(2023{\natexlab{a}})Liu, Lin, Li, Wang, Yacoob, and Wang]{liu2023mitigating}
Fuxiao Liu, Kevin Lin, Linjie Li, Jianfeng Wang, Yaser Yacoob, and Lijuan Wang.
\newblock Mitigating hallucination in large multi-modal models via robust instruction tuning.
\newblock In \emph{The Twelfth International Conference on Learning Representations}, 2023{\natexlab{a}}.

\bibitem[Liu et~al.(2024{\natexlab{b}})Liu, Li, Li, and Lee]{liu2024improved}
Haotian Liu, Chunyuan Li, Yuheng Li, and Yong~Jae Lee.
\newblock Improved baselines with visual instruction tuning.
\newblock In \emph{Proceedings of the IEEE/CVF Conference on Computer Vision and Pattern Recognition}, pages 26296--26306, 2024{\natexlab{b}}.

\bibitem[Liu et~al.(2024{\natexlab{c}})Liu, Li, Li, Li, Zhang, Shen, and Lee]{liu2024llavanext}
Haotian Liu, Chunyuan Li, Yuheng Li, Bo Li, Yuanhan Zhang, Sheng Shen, and Yong~Jae Lee.
\newblock Llava-next: Improved reasoning, ocr, and world knowledge, 2024{\natexlab{c}}.

\bibitem[Liu et~al.(2024{\natexlab{d}})Liu, Li, Wu, and Lee]{liu2024visual}
Haotian Liu, Chunyuan Li, Qingyang Wu, and Yong~Jae Lee.
\newblock Visual instruction tuning.
\newblock \emph{Advances in neural information processing systems}, 36, 2024{\natexlab{d}}.

\bibitem[Liu et~al.(2023{\natexlab{b}})Liu, Zeng, Ren, Li, Zhang, Yang, Li, Yang, Su, Zhu, et~al.]{liu2023grounding}
Shilong Liu, Zhaoyang Zeng, Tianhe Ren, Feng Li, Hao Zhang, Jie Yang, Chunyuan Li, Jianwei Yang, Hang Su, Jun Zhu, et~al.
\newblock Grounding dino: Marrying dino with grounded pre-training for open-set object detection.
\newblock \emph{arXiv preprint arXiv:2303.05499}, 2023{\natexlab{b}}.

\bibitem[Liu et~al.(2023{\natexlab{c}})Liu, Duan, Zhang, Li, Zhang, Zhao, Yuan, Wang, He, Liu, et~al.]{liu2023mmbench}
Yuan Liu, Haodong Duan, Yuanhan Zhang, Bo Li, Songyang Zhang, Wangbo Zhao, Yike Yuan, Jiaqi Wang, Conghui He, Ziwei Liu, et~al.
\newblock Mmbench: Is your multi-modal model an all-around player?
\newblock \emph{arXiv preprint arXiv:2307.06281}, 2023{\natexlab{c}}.

\bibitem[Liu et~al.(2025)Liu, Duan, Zhang, Li, Zhang, Zhao, Yuan, Wang, He, Liu, et~al.]{liu2025mmbench}
Yuan Liu, Haodong Duan, Yuanhan Zhang, Bo Li, Songyang Zhang, Wangbo Zhao, Yike Yuan, Jiaqi Wang, Conghui He, Ziwei Liu, et~al.
\newblock Mmbench: Is your multi-modal model an all-around player?
\newblock In \emph{European Conference on Computer Vision}, pages 216--233. Springer, 2025.

\bibitem[Lu et~al.(2024)Lu, Liu, Zhang, Wang, Dong, Liu, Sun, Ren, Li, Sun, et~al.]{lu2024deepseek}
Haoyu Lu, Wen Liu, Bo Zhang, Bingxuan Wang, Kai Dong, Bo Liu, Jingxiang Sun, Tongzheng Ren, Zhuoshu Li, Yaofeng Sun, et~al.
\newblock Deepseek-vl: towards real-world vision-language understanding.
\newblock \emph{arXiv preprint arXiv:2403.05525}, 2024.

\bibitem[Lu et~al.(2022)Lu, Mishra, Xia, Qiu, Chang, Zhu, Tafjord, Clark, and Kalyan]{lu2022learn}
Pan Lu, Swaroop Mishra, Tanglin Xia, Liang Qiu, Kai-Wei Chang, Song-Chun Zhu, Oyvind Tafjord, Peter Clark, and Ashwin Kalyan.
\newblock Learn to explain: Multimodal reasoning via thought chains for science question answering.
\newblock \emph{Advances in Neural Information Processing Systems}, 35:\penalty0 2507--2521, 2022.

\bibitem[Openai(2024)]{gpt_4o}
Openai.
\newblock Hello gpt-4o.
\newblock \url{https://openai.com/index/hello-gpt-4o/}, 2024.

\bibitem[Peebles and Xie(2023)]{peebles2023scalable}
William Peebles and Saining Xie.
\newblock Scalable diffusion models with transformers.
\newblock In \emph{Proceedings of the IEEE/CVF International Conference on Computer Vision}, pages 4195--4205, 2023.

\bibitem[Rafailov et~al.(2024)Rafailov, Sharma, Mitchell, Manning, Ermon, and Finn]{rafailov2024direct}
Rafael Rafailov, Archit Sharma, Eric Mitchell, Christopher~D Manning, Stefano Ermon, and Chelsea Finn.
\newblock Direct preference optimization: Your language model is secretly a reward model.
\newblock \emph{Advances in Neural Information Processing Systems}, 36, 2024.

\bibitem[Rohrbach et~al.(2018)Rohrbach, Hendricks, Burns, Darrell, and Saenko]{rohrbach2018object}
Anna Rohrbach, Lisa~Anne Hendricks, Kaylee Burns, Trevor Darrell, and Kate Saenko.
\newblock Object hallucination in image captioning.
\newblock \emph{arXiv preprint arXiv:1809.02156}, 2018.

\bibitem[Saikh et~al.(2022)Saikh, Ghosal, Mittal, Ekbal, and Bhattacharyya]{saikh2022scienceqa}
Tanik Saikh, Tirthankar Ghosal, Amish Mittal, Asif Ekbal, and Pushpak Bhattacharyya.
\newblock Scienceqa: A novel resource for question answering on scholarly articles.
\newblock \emph{International Journal on Digital Libraries}, 23\penalty0 (3):\penalty0 289--301, 2022.

\bibitem[Schulman et~al.(2017)Schulman, Wolski, Dhariwal, Radford, and Klimov]{schulman2017proximal}
John Schulman, Filip Wolski, Prafulla Dhariwal, Alec Radford, and Oleg Klimov.
\newblock Proximal policy optimization algorithms.
\newblock \emph{arXiv preprint arXiv:1707.06347}, 2017.

\bibitem[Sharma et~al.(2018)Sharma, Ding, Goodman, and Soricut]{sharma2018conceptual}
Piyush Sharma, Nan Ding, Sebastian Goodman, and Radu Soricut.
\newblock Conceptual captions: A cleaned, hypernymed, image alt-text dataset for automatic image captioning.
\newblock In \emph{Proceedings of the 56th Annual Meeting of the Association for Computational Linguistics (Volume 1: Long Papers)}, pages 2556--2565, 2018.

\bibitem[Tang et~al.()Tang, Huang, Liu, Sun, Yang, and Lim]{tangintervening}
Feilong Tang, Zile Huang, Chengzhi Liu, Qiang Sun, Harry Yang, and Ser-Nam Lim.
\newblock Intervening anchor token: Decoding strategy in alleviating hallucinations for mllms.
\newblock In \emph{The Thirteenth International Conference on Learning Representations}.

\bibitem[Touvron et~al.(2023)Touvron, Lavril, Izacard, Martinet, Lachaux, Lacroix, Rozi{\`e}re, Goyal, Hambro, Azhar, et~al.]{touvron2023llama}
Hugo Touvron, Thibaut Lavril, Gautier Izacard, Xavier Martinet, Marie-Anne Lachaux, Timoth{\'e}e Lacroix, Baptiste Rozi{\`e}re, Naman Goyal, Eric Hambro, Faisal Azhar, et~al.
\newblock Llama: Open and efficient foundation language models.
\newblock \emph{arXiv preprint arXiv:2302.13971}, 2023.

\bibitem[Wang et~al.(2024{\natexlab{a}})Wang, Bai, Tan, Wang, Fan, Bai, Chen, Liu, Wang, Ge, et~al.]{wang2024qwen2}
Peng Wang, Shuai Bai, Sinan Tan, Shijie Wang, Zhihao Fan, Jinze Bai, Keqin Chen, Xuejing Liu, Jialin Wang, Wenbin Ge, et~al.
\newblock Qwen2-vl: Enhancing vision-language model's perception of the world at any resolution.
\newblock \emph{arXiv preprint arXiv:2409.12191}, 2024{\natexlab{a}}.

\bibitem[Wang et~al.(2024{\natexlab{b}})Wang, Pan, Ding, and Biemann]{wang2024mitigating}
Xintong Wang, Jingheng Pan, Liang Ding, and Chris Biemann.
\newblock Mitigating hallucinations in large vision-language models with instruction contrastive decoding.
\newblock \emph{arXiv preprint arXiv:2403.18715}, 2024{\natexlab{b}}.

\bibitem[Wang et~al.(2024{\natexlab{c}})Wang, Bi, Pentyala, Ramnath, Chaudhuri, Mehrotra, Mao, Asur, et~al.]{wang2024comprehensive}
Zhichao Wang, Bin Bi, Shiva~Kumar Pentyala, Kiran Ramnath, Sougata Chaudhuri, Shubham Mehrotra, Xiang-Bo Mao, Sitaram Asur, et~al.
\newblock A comprehensive survey of llm alignment techniques: Rlhf, rlaif, ppo, dpo and more.
\newblock \emph{arXiv preprint arXiv:2407.16216}, 2024{\natexlab{c}}.

\bibitem[Xiao et~al.(2024)Xiao, Huang, Gan, He, Li, Yu, Jiang, Wu, and Zhu]{xiao2024detecting}
Wenyi Xiao, Ziwei Huang, Leilei Gan, Wanggui He, Haoyuan Li, Zhelun Yu, Hao Jiang, Fei Wu, and Linchao Zhu.
\newblock Detecting and mitigating hallucination in large vision language models via fine-grained ai feedback.
\newblock \emph{arXiv preprint arXiv:2404.14233}, 2024.

\bibitem[Xu et~al.(2024)Xu, Wu, Diao, Liu, Wang, Chen, and Gao]{xu2024sayself}
Tianyang Xu, Shujin Wu, Shizhe Diao, Xiaoze Liu, Xingyao Wang, Yangyi Chen, and Jing Gao.
\newblock Sayself: Teaching llms to express confidence with self-reflective rationales.
\newblock \emph{arXiv preprint arXiv:2405.20974}, 2024.

\bibitem[Xue et~al.(2025)Xue, Tang, Hu, Liu, Huang, Li, Liu, Xu, Zhang, Feng, et~al.]{xue2025mmrc}
Haochen Xue, Feilong Tang, Ming Hu, Yexin Liu, Qidong Huang, Yulong Li, Chengzhi Liu, Zhongxing Xu, Chong Zhang, Chun-Mei Feng, et~al.
\newblock Mmrc: A large-scale benchmark for understanding multimodal large language model in real-world conversation.
\newblock \emph{arXiv preprint arXiv:2502.11903}, 2025.

\bibitem[Yang et~al.(2024)Yang, Yang, Hui, Zheng, Yu, Zhou, Li, Li, Liu, Huang, et~al.]{yang2024qwen2}
An Yang, Baosong Yang, Binyuan Hui, Bo Zheng, Bowen Yu, Chang Zhou, Chengpeng Li, Chengyuan Li, Dayiheng Liu, Fei Huang, et~al.
\newblock Qwen2 technical report.
\newblock \emph{arXiv preprint arXiv:2407.10671}, 2024.

\bibitem[Yao et~al.(2024)Yao, Yu, Zhang, Wang, Cui, Zhu, Cai, Li, Zhao, He, et~al.]{yao2024minicpm}
Yuan Yao, Tianyu Yu, Ao Zhang, Chongyi Wang, Junbo Cui, Hongji Zhu, Tianchi Cai, Haoyu Li, Weilin Zhao, Zhihui He, et~al.
\newblock Minicpm-v: A gpt-4v level mllm on your phone.
\newblock \emph{arXiv preprint arXiv:2408.01800}, 2024.

\bibitem[Yin et~al.(2023)Yin, Fu, Zhao, Xu, Wang, Sui, Shen, Li, Sun, and Chen]{yin2023woodpecker}
Shukang Yin, Chaoyou Fu, Sirui Zhao, Tong Xu, Hao Wang, Dianbo Sui, Yunhang Shen, Ke Li, Xing Sun, and Enhong Chen.
\newblock Woodpecker: Hallucination correction for multimodal large language models.
\newblock \emph{arXiv preprint arXiv:2310.16045}, 2023.

\bibitem[Yu et~al.(2024)Yu, Li, Wei, Pang, Ye, Qin, Tang, Tian, and Zhuang]{yu2024hallucidoctor}
Qifan Yu, Juncheng Li, Longhui Wei, Liang Pang, Wentao Ye, Bosheng Qin, Siliang Tang, Qi Tian, and Yueting Zhuang.
\newblock Hallucidoctor: Mitigating hallucinatory toxicity in visual instruction data.
\newblock In \emph{Proceedings of the IEEE/CVF Conference on Computer Vision and Pattern Recognition}, pages 12944--12953, 2024.

\bibitem[Yu et~al.(2023)Yu, Yang, Li, Wang, Lin, Liu, Wang, and Wang]{yu2023mm}
Weihao Yu, Zhengyuan Yang, Linjie Li, Jianfeng Wang, Kevin Lin, Zicheng Liu, Xinchao Wang, and Lijuan Wang.
\newblock Mm-vet: Evaluating large multimodal models for integrated capabilities.
\newblock \emph{arXiv preprint arXiv:2308.02490}, 2023.

\bibitem[Zhang et~al.(2024)Zhang, Yu, Wen, Wang, Zhang, Wang, Jin, and Tan]{zhang2024debiasing}
Yi-Fan Zhang, Weichen Yu, Qingsong Wen, Xue Wang, Zhang Zhang, Liang Wang, Rong Jin, and Tieniu Tan.
\newblock Debiasing large visual language models.
\newblock \emph{arXiv preprint arXiv:2403.05262}, 2024.

\bibitem[Zhao et~al.(2023)Zhao, Wang, Ouyang, Dong, Wang, and He]{zhao2023beyond}
Zhiyuan Zhao, Bin Wang, Linke Ouyang, Xiaoyi Dong, Jiaqi Wang, and Conghui He.
\newblock Beyond hallucinations: Enhancing lvlms through hallucination-aware direct preference optimization.
\newblock \emph{arXiv preprint arXiv:2311.16839}, 2023.

\bibitem[Zhu et~al.(2024)Zhu, Zhao, Ge, and Zhang]{zhu2024self}
Ke Zhu, Liang Zhao, Zheng Ge, and Xiangyu Zhang.
\newblock Self-supervised visual preference alignment.
\newblock \emph{arXiv preprint arXiv:2404.10501}, 2024.

\end{thebibliography}
}


\end{document}